\documentclass[11pt,english]{article}
\usepackage[sc]{mathpazo}
\usepackage{geometry}
\usepackage{color}
\usepackage{array}
\usepackage{bbding}
\usepackage{multirow}
\usepackage[fleqn]{amsmath}
\usepackage{amssymb}
\usepackage{amsthm}
\usepackage{graphicx}
\usepackage{subfig}
\usepackage{natbib}
\usepackage{lscape}
\usepackage[hidelinks]{hyperref}
\usepackage{comment}
\usepackage{bm}
\usepackage{float}
\usepackage{longtable}
\usepackage{mathtools}
\usepackage{chngcntr}
\usepackage{authblk}
\makeatletter
\allowdisplaybreaks

\@ifundefined{showcaptionsetup}{}{%
 \PassOptionsToPackage{caption=false}{subfig}}
\usepackage{subfig}
\makeatother

\usepackage{babel}
\usepackage{enumitem}

\begin{document}\sloppy

% \title{Multi-scale Graph Wavelet-based Temporal Convolution Neural Networks for Traffic Prediction for Complex Transportation Networks}
% \title{Learning Multiple Levels of Spatial-temporal Information in Traffic State Prediction for Complex Transportation Networks: A Multi-scale Graph Wavelet-based Approach}
\title{Traffic Prediction considering Multiple Levels of Spatial-temporal Information: A Multi-scale Graph Wavelet-based Approach}
\author{Zilin Bian \textsuperscript{a}\hspace{3em} Jingqin Gao\textsuperscript{a}\hspace{3em} Kaan Ozbay\textsuperscript{b}\hspace{3em} Zhenning Li\textsuperscript{c}\hspace{3em}}
\affil{\small\emph{\textsuperscript{a}Department of Civil and Urban Engineering, Tandon School of Engineering, New York University, 6 MetroTech Center, 4th Floor, Brooklyn, NY, 11201, USA}\normalsize}
\affil{\small\emph{\textsuperscript{b}Department of Civil and Urban Engineering, Tandon School of Engineering, New York University, 6 MetroTech Center, 4th Floor, Brooklyn, NY, 11201, USA}\normalsize}
\affil{\small\emph{\textsuperscript{c}State Key Laboratory of Internet of Things for Smart City, University of Macau, N21 5025a, Avenida da Universidade, Taipa, Macau, People's Republic of China}\normalsize}

%\date{\today}
\maketitle

\begin{abstract}
\noindent Although traffic prediction has been receiving considerable attention with a number of successes in the context of intelligent transportation systems, the prediction of traffic states over a complex transportation network that contains different road types has remained a challenge. This study proposes a multi-scale graph wavelet temporal convolution network (MSGWTCN) to predict the traffic states in complex transportation networks. Specifically, a multi-scale spatial block is designed to simultaneously capture the spatial information at different levels, and the gated temporal convolution network is employed to extract the temporal dependencies of the data. The model jointly learns to mount multiple levels of the spatial interactions by stacking graph wavelets with different scales. Two real-world datasets are used in this study to investigate the model performance, including a highway network in Seattle and a dense road network of Manhattan in New York City. Experiment results show that the proposed model outperforms other baseline models. Furthermore,  different scales of graph wavelets are found to be effective in extracting local, intermediate and global information at the same time and thus enable the model to learn a complex transportation network topology with various types of road segments. By carefully customizing the scales of wavelets, the model is able to improve the prediction performance and better adapt to different network configurations.\par 

\hfill\break%
\end{abstract}

\section{Introduction}\label{sc:introduction}
Predictability of time-varying traffic states in complex transportation networks is a primary concern for the transportation systems in smart cities. Accurate and timely prediction can help the transportation agencies, as well as road users, to better manage and respond to the recurrent and non-recurrent traffic congestion. \par

Besides the traditional statistic models (\citealp{kumar2015short}, \citealp{chen2011short}), deep learning approaches, including recurrent neural network (RNN) and convolutional neural network (CNN), are often used to predict traffic states (\citealp{ma2015long}, \citealp{zhou2017recurrent}). % ...data-driven methods that utilize increasingly available large-scale, real-time traffic data to employ a variety of machine learning algorithms. As artificial intelligence (AI) becomes ubiquitous, many deep learning models, including recurrent neural network (RNN), convolutional neural network (CNN), and their variants have drawn increasing attention in the field . 
Lately, the graph neural network (GNN) and its variants, such as graph convolutional network (GCN), have proven successful in dealing with complex structures of the traffic networks, as they can capture the topology and dependence of graphs via message-passing between the roadway segments and intersections (\citealp{zhou2020graph}). One limitation of the traditional GCN model is that it assumes uniform interactions between the primary road link and its neighboring road segments (\citealp{yu2020forecasting}) which is unrealistic in the real world. Researchers have attempted to resolve this issue by introducing the connection intensities via methods like attention mechanisms (\citealp{do2019effective, liu2020dynamic}). However, major drawbacks related to this approach still exist. For example, spatial features are centrally localized at the primary graph node (\citealp{cui2019traffic}) and confined to the same finite number of hops in neighboring nodes (e.g., k-hop neighborhood). This may not reflect reality, since the scale of the traffic propagation may vary due to the external factors such as accidents, severe weather, construction and so on.\par 
To adequately capture the localized features, a graph wavelet neural network (GWNN) (\citealp{xu2019graph}) that adopts the graph wavelet transform in the spectral GCN was introduced recently for node classification. It has good localization property in the vertex domain, and the diffusion range of features can be fine-tuned in a continuous manner, making it more flexible in capturing the node features (\citealp{tremblay2014graph}).\par

However, current practices still face two challenges. Firstly, using only localized features are not sufficient to completely represent the traffic states. In fact, the traffic congestion is influenced not only by the traffic states of its neighboring road segments, but also by other key regions that are located a certain distance away from the congested site. This is especially true for the traffic networks with limited access roads, such as two regions connected by several bridges or tunnels. To be able to accurately predict traffic states, it is necessary to aggregate multilevel structural information during the feature extraction process. \par

Secondly, a strong temporal correlation usually exists in the traffic data, where the current traffic states are likely to be impacted by the previous traffic conditions. As a result, some traffic predictors that combined with an RNN structure have been used to resolve the problem of temporal prediction. For example, the long short-term memory (LSTM) (\citealp{ma2015long}) and gated structure like gated recurrent unit (GRU) (\citealp{ cho2014properties}) have been used to mine the temporal features when making predictions (\citealp{ma2015large}; \citealp{ yu2017spatiotemporal}; \citealp{do2019effective}). However,  RNN-based models handle the time-series data in a recursive manner and meet the gradient explosion issues. As opposed to RNN-based models, the gated temporal convolution network (TCN) (\citealp{oord2016wavenet}) handles the sequences in a non-recursive manner and alleviates the gradient explosion problems. It is also able to facilitate parallel computation which handles the temporal units in a cheaper way. %Therefore, this paper adopted gated TCN instead of RNN-based models to accommodate the temporal correlations. 
\par 

In addition, the traffic prediction in complex transportation networks (e.g., consists of various types of road segments such as highways, ramps, arterials, one-way streets) has remained intractable when using GCN-based approaches. Can a GCN-based model properly extract the system's spatial features, given this variety? Moreover, because such complexity also varies across different transportation network structures, a GCN-based model may not be sensitive enough to such structure difference and therefore unable to achieve viable traffic prediction performance. \par 

In this study, we proposes a novel, multi-scale graph wavelet-based neural network to mine the spatial features among all roadway links and combines it with the gated temporal convolution network to learn the temporal features of traffic congestion evolution. Therefore, the model is able to learn multiple levels of spatial and temporal features at the same time. The contributions of the paper can be summarized as follows:\par
\begin{itemize}
        \item We developed a novel, multi-scale hybrid model that combines multiple graph wavelets with the gated TCN to predict traffic states in a complex transportation network. The proposed model incorporates a multi-scale structure to learn different levels of structural information of traffic networks.
        \item The proposed architecture was validated via the experiments using two real-world traffic datasets, one in Seattle and one in NYC. Both datasets verified that our proposed model outperforms other state-of-the-art models in the prediction accuracy. \par
        \item The evaluation of the graph wavelet weights proved that the multi-scale setting of the model is able to improve the prediction performance by capturing the local, intermediate and global information of spatial dependencies simultaneously. Different scales of graph wavelets can extract the information from multiple types of road links and thus enable the model to learn a complex transportation network topology with various types of road segments.
        \item Ablation experiment revealed that the transportation network structures are sensitive to different multi-scale graph wavelet settings. The results provide guidance on the selection of the scales for different network configurations and better prediction can be obtained by carefully customizing the scales of wavelets. %Large-scale, highway-only transportation networks can achieve the best prediction performance by combining small, medium and large scales of graph wavelets, while combining large scales of graph wavelets only performed best for small-scale network with mainly urban streets. \par  
\end{itemize}

In the rest of the paper, we first discuss existing studies of traffic prediction, graph neural networks and multi-scale feature learning in Section \ref{sc:literature}. Second, we propose our methodology of combining the multi-scale graph wavelets with gated temporal convolution network (TCN) layers and introduce some graph wavelet related concepts and some commonly used neighborhood aggregation methods in Section \ref{sc:methodology}. Third, we provide the experiment settings and compared our proposed model with other prevailing baseline models in terms of the popular evaluation metrics in Section \ref{sc:experiment}. Fourth, we extract the learned weight parameters to investigate the effect of multi-scale graph wavelets in Section \ref{sc:result}. We show the capability of interpreting our model using various forms of visualization. Finally, we conclude with our findings and future steps in Section \ref{sc:conclusion}.

\section{Literature Review}\label{sc:literature}
Various approaches have been introduced to model future traffic conditions based on previously observed time-series. Classical statistical inference models, such as the autoregressive integrated moving average (ARIMA) model (\citealp{van1996combining, williams2001multivariate}), Kalman filter (\citealp{okutani1984dynamic}), as well as Bayesian inference methods (\citealp{sun2006bayesian, ghosh2007bayesian, li2019bayesian}), were widely used in recent decades. More recent studies are concentrated on using the hybrid structures of Neural Networks (NNs) and optimization techniques like genetic algorithms or wavelets to predict both spatial and temporal dependencies of the traffic networks (\citealp{vlahogianni2014short}). These models usually outperform classic models, particularly when modeling multi-dimensional or large-scale datasets.\par

\subsection{Graph-based deep learning models in Transportation}
The power of GNN in modeling the dependence of graph through information exchanged between nodes enables wide application to deal with the non-linear, structured, real-world data. Recently, different types of GNNs such as the graph convolutional network (GCN) (\citealp{yu2020forecasting}), the graph attention network (GAT) (\citealp{zhang2019spatial}), and the graph recurrent network (GRN) (\citealp{do2019effective}) have demonstrated convincing performances in the transportation domain. For example, Zhang et al. (\citealp{zhang2019spatial}) integrated spatio-temporal variables into a GCN for the multistep traffic speed prediction. An attention mechanism with a sequence-to-sequence model framework was used to capture the temporal heterogeneity of network traffic and overcome the multi-step prediction challenges.\par

Using continuous feedback between time steps, various types of neural networks dealing with the temporal dependencies have also been proposed for the traffic flow prediction problems and applied with GNNs, such as LSTM (\citealp{cui2019traffic}; \citealp{bogaerts2020graph}), GRU (\citealp{do2019effective}), TCN (\citealp{zhao2019t}) and Residual Neural Networks (\citealp{chen2019gated}). The gated mechanisms (\citealp{zhao2019t}; \citealp{chen2019gated}; \citealp{cui2020learning}) are often used to memorize the long-term information and overcome the vanishing gradient problem of RNN. \par

\subsection{Spectral Graph Wavelet Transform}
The spectral graph wavelet transforms firstly introduced by Hammond et al. (\citealp{hammond2011wavelets}) have been proven to exhibit the good localization properties in the fine scale limit and showed their potential in a variety of different problem domains, such as brain imaging. Combining the spectral graph wavelet transform with a neural network, Xu et al. (\citealp{xu2019graph}) developed a novel GCN, graph wavelet neural network (GWNN), to detach the feature transformation from the convolution and introduce the graph wavelet transform to solve semi-supervised classification problems. Different than previous spectral GCNs based on graph Fourier transform, the graph wavelets are sparse and provide good localization and interpretability for the graph convolution. The graph wavelets can be efficiently approximated through the Chebyshev polynomials method and thus avoid the use of eigendecomposition, which requires high computation effort (\citealp{xu2019graph}). \par 

Although the GWNN has been successfully applied to domains like the breast cancer diagnose (\citealp{zhang2020ms}), node classification (\citealp{xu2019graph}) and imaging classification (\citealp{chang2020spectral}), to the best of our knowledge, there are only a few papers adopting GWNN for network traffic prediction. Cui et al.(\citealp{cui2020learning}) proposed a graph wavelet gated recurrent (GWGR) neural network to capture the complex spatial-temporal features for the large-scale transportation data. Their results showed that graph wavelet-based approach is effective in extracting localized spatial features for the traffic speed prediction. In addition, a gated RNN structure was employed to learn the temporal dependencies. However, these experiments were performed only at one fixed scale that still confines a finite range of neighborhoods and may miss important traffic information from other network scales. Thus, there is a need to perform multi-scale analysis to account for different levels of neighborhoods. Recent success in applying the GWNN model in breast cancer diagnose (\citealp{zhang2020ms}) with the multi-scale contextual interactions in the whole pathological slide by aggregating features at different scales showed the potential power of employing the multi-scale information. 

\subsection{Multi-Scale Feature Learning in Transportation}
The multi-scale feature learning has proven effective in performing many tasks in the transportation domain, including the traffic state prediction (\citealp{zang2018long}; \citealp{dai2019deeptrend}), traffic demand prediction (\citealp{chu2019deep}), sign recognition (\citealp{sermanet2011traffic}), mode classification (\citealp{zhang2019classifying}), and so on. Complex features can be extracted by learning the multi-scale inputs from both spatial and temporal domains. A multi-scale feature learning model (\citealp{zang2018long}) based on the convolutional LSTM and CNN was proposed to predict the long-term traffic speed for elevated highways. Different sizes of convolutional LSTMs, CNNs and speed matrices generated from the historical data were used as input for learning features of three time scales to predict the traffic speed of a certain day (\citealp{zang2018long}). Similarly, a multi-scale spatiotemporal traffic prediction model, named MT-STNets (\citealp{wang2021mt}), was developed to deal with both the fine- and coarse-grained traffic data based on traditional GCN. However, this model only combined the nearby nodes to define a global neighborhood and required empirical analysis to cluster coarse-grained regions. \par
Motivated by the power of the graph wavelet transform and the need to perform the traffic prediction on different scales, this study proposes a multi-scale model based on the graph wavelet and TCNs to predict the network-wide traffic state that extracts localized, intermediate and global information simultaneously. 
\par

\section{Methodology}\label{sc:methodology}
In this section, we propose a new data-driven framework by incorporating both spatial and temporal blocks to predict the traffic states. Then we introduce the details of each block in the framework and provide the mathematical formulas. Specifically, we show how to form the temporal block using a gated TCN and the mathematical processes to address the spatial dependencies through the aggregation of multiple scales of graph wavelets.\par

\subsection{Notations and problem definition}\label{ssc:notation}
Consider a general transportation network as a directed graph $\mathcal{G = (V,E)}$, where $\mathcal{V}$ and $\mathcal{E}$ are the sets of nodes and edges in the network, respectively. We set $v_j \in \mathcal{V}$ as node $i$ and $(v_i, v_j)\in \mathcal{E}$ as the edge linking node $v_i$ and $v_j$.  Let $\mathbf{A} \in \mathbb{R}^{N\times N}$ denote the adjacency matrix of graph $\mathcal{G}$, $\mathbf{A}$ represents the connectivity of nodes and $N$ represents the total number of nodes in the graph. $A_{ij}=1$ if node $v_i$ and $v_j$ are connected, otherwise $A_{ij}=0$. We denote $\mathbf{L}$ as the graph Laplacian matrix, $\mathbf{L}$ is usually defined in combinatorial form $\mathbf{L= D-A}$ or normalized form $\mathbf{L}=I_N-\mathbf{D}^{-1/2}\mathbf{A}\mathbf{D}^{1/2}$, where $I_N\in \mathbb{R}^{N\times N}$ is identity matrix and $\mathbf{D} \in \mathbb{R}^{N\times N}$ is the diagonal degree matrix. $\mathbf{L}$ can be decomposed using Laplacian eigenvectors $U$: $\mathbf{L=D-A}=U \Lambda U^{-1}$, where $\Lambda = diag(\lambda_1, ..., \lambda_N)$ is the diagonal matrix consisting of eigenvalues $\lambda$ of $\mathbf{L}$. \par 
Define $X_t\in \mathbb{R}^{N\times C}$ as the traffic states at time step $t$, where $C$ denotes the number of traffic states (e.g., traffic volume, traffic speed, etc) on each node in $\mathcal{G}$. In this study, we denote $\mathcal{G}$ as the transportation network and each node in $\mathcal{G}$ represents a traffic sensor.
Our problem is to learn a mapping function $F$ given the observations at $N$ nodes of historical $P$ time steps, the observations are denote as $X = (X_{t_1}, X_{t_2}, ..., X_{t_P}) \in \mathbb{R}^{P\times N\times C}$. The mapping function $F$ is represented as follows:

\begin{align}
    \hat{Y} = F(X) \label{e1}
\end{align}

In this study, the mapping function $F$ is used to predict traffic states $\hat{Y} = (\hat{X}_{t_{P+1}}, \hat{X}_{t_{P+2}}, ..., \hat{X}_{t_{P+T}}) \in \mathbb{R}^{T\times N\times C}$ in the next $T$ time steps, given $P$ previous traffic data $X$.

\subsection{Framework of multi-scale graph wavelet-based temporal convolution networks}
\begin{figure}[hbt]
    \centering
    \includegraphics[width=0.5\textwidth]{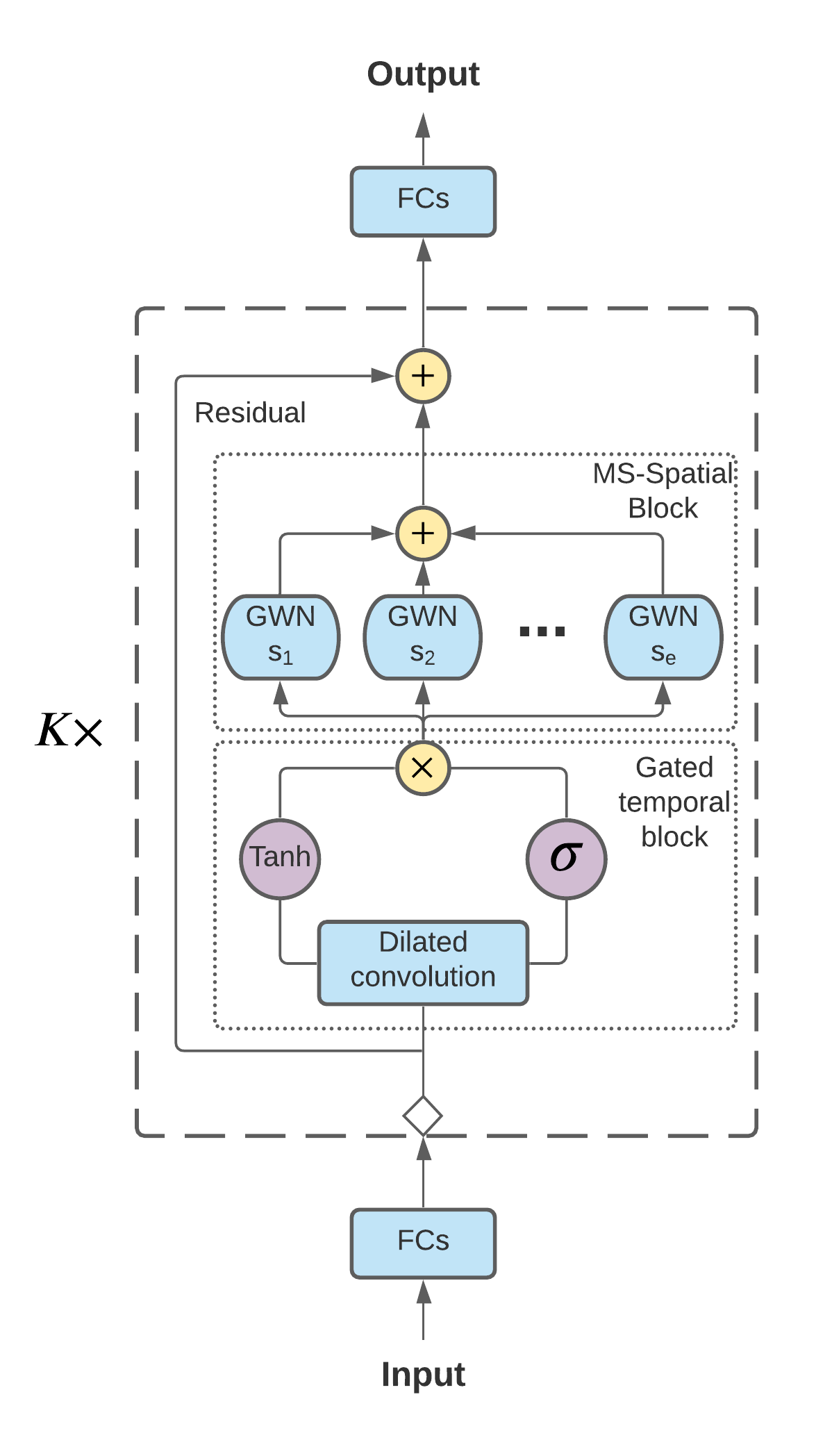}
    \caption{Framework of multi-scale graph wavelet-based gated temporal convolution network}
    \label{fig:multi-scale model framework}
\end{figure}

In order to obtain multiple levels of spatial-temporal features for each node in graph $\mathcal{G}$, we propose a multi-scale graph wavelet-based temporal convolution network (MSGWTCN). MSGWTCN consists of $K$ stacked layers, and each layer contains one temporal block and one spatial block with residual connections. Before and after the spatial and temporal blocks, fully-connected layers are added to transform the input and output information. To be specific, each layer contains one multi-scale spatial block (MS-spatial block) to capture the spatial features and one gated temporal block to capture the temporal information of node features. The MS-spatial block employs several graph wavelet nets (GWNs) and aggregates them to extract the spatial information at multiple levels. The gated temporal block adopts gated TCN to learn the temporal correlations. The framework is proposed in Fig.\ref{fig:multi-scale model framework}. The details of each block in this framework will be introduced next.

\subsection{Gated temporal block}\label{ssc:gtcn}
\begin{figure}[htp]
    \centering
    \includegraphics[width=0.9\textwidth]{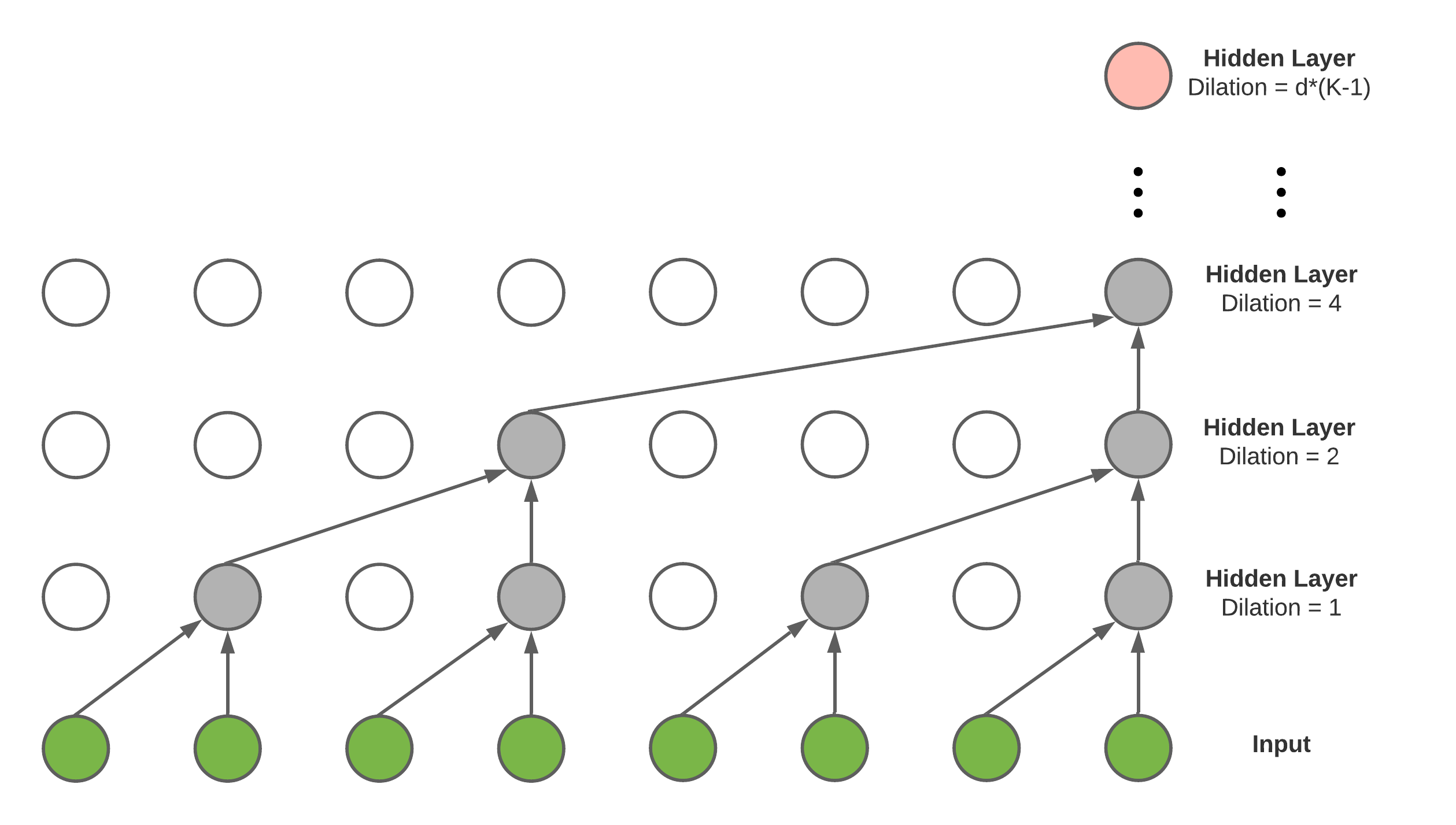}
    \caption{Multiple layers of causal dilation convolution}
    \label{fig:dilationconvolution}
\end{figure}

At the gate temporal block, We adopt gated TCN to capture the temporal dependencies of the formulated prediction problem. Instead of using TCN only, we insert gating units into TCN since they work well in controlling information flow in $K$ stacked layers (\citealp{dauphin2017language}). The gated TCN is composed of dilated causal convolution and gated activation units. The dilated causal convolution does not propagate gradients following a temporal direction path with recursive connections, which avoids the problem of gradient explosion, a common drawback of recurrent neural network models. Moreover, the settings of dilated causal convolution (as shown in Fig. \ref{fig:dilationconvolution}) can be made in a parallel structure which alleviates the computation effort via parallel computing.\par
Given a 1-D sequence input signal $\mathbf{x} \in \mathbb{R}^T$ and a convolution kernel $\mathbf{f} \in \mathbb{R}^K$, the dilated causal convolution at time $t$ is as follows:
\begin{align}
    \mathbf{x} * \mathbf{f} (t) = \sum_{i=0}^{K-1} \mathbf{f}_i \mathbf{x}(t-d\times i) \label{e8}
\end{align}
where $d$ is the dilation operator which controls how many steps to dilate /skip along the temporal direction of previous time steps. \par
By increasing the order of the dilation causal convolution as shown in Eq.\ref{e8}, the receptive field of the model grows exponentially, which enables the model to capture the global contextual information of sequences without requiring a large computational effort. 

Beyond the dilated causal convolution, two gating units, tanh and sigmoid, are inserted to better facilitate the propagation process among the stacked causal dilation convolution layers. The gating activation is as follows:
\begin{align}
    h = tanh(\mathbf{W_1}*\mathbf{x}) \odot \sigma(\mathbf{W_2} * \mathbf{x})\label{e9}
\end{align}
where $h$ denotes the hidden features, $\sigma$ denotes the sigmoid activation unit, $\tanh$ denotes the tanh activation unit, and $\mathbf{W_1}, \mathbf{W_2}$ denote the learnable parameters for convolution filters. \par
In summary, we employ the causal dilated convolution with two gating units in the gated temporal block to learn the temporal features. The use of gating units can better facilitate the weight transition process. By stacking multiple layers with different dilated factors, the gated temporal block can capture the temporal dependencies in different levels. To be specific, the the gated temporal block at the bottom layer extracts the short-term temporal information while the long-term information can be captured at the top layer. The gated temporal block is further accompanied with spatial information via the MS-spatial block. We will show the mathematical details of MS-spatial block next.

\subsection{Graph Fourier transform and graph wavelet transform}\label{ssc:gftgwt}
The graph Fourier transform provides a way to project the graph signal into the vertex domain or spectral domain. Similarly, graph wavelet transform projects the graph signal into the spectral domain. The graph Fourier transform requires the eigen-decomposition of Laplacian matrix, which commonly requires $\mathbb{O}(n^3)$ computation complexity. Proposed by (\citealp{hammond2011wavelets}), the graph wavelets can be obtained using the Chebyshev polynomial approximation method, which reduces the computation complexity from $\mathbb{O}(n^3)$ to $\mathbb{O}(n\times p)$, where $p$ denotes the approximation order.\par 

% The major difference between the graph Fourier transform and the graph wavelet transform is that the graph Fourier transform takes the eigenvectors of Laplacian matrix as spectral basis $U$ while the graph wavelet transform employs wavelets. 
Denoting the eigenvectors as $U$ in the Fourier basis, the graph Fourier transform of a graph signal $x \in \mathbb{R}^N$ on graph $\mathcal{G}$ can be defined using $U$ as $\hat{x} = U^T x$, the inverse graph Fourier transform is $x=U\hat{x}$, where $x$ denotes the graph signal in $\mathcal{G}$. For graph Fourier transform, the convolution operator, denoted as $*_{\mathcal{G}}$ on graph $\mathcal{G}$ can be defined in a generalized form:
\begin{align}
    x*_{\mathcal{G}} y=U((U^T y)\odot (U^T x)) \label{e2}
\end{align}
where $\odot$ is the element-wise Hadamard product, $y$ is the convolutional kernel using graph Fourier transform. \par
Besides the generalized form, the graph convolution can also be written in a different form by re-writing the Hadamard product in a matrix form. By replacing the $U^T y$ with a diagonal matrix $g(\Lambda_{\theta})$, Eq.\ref{e2} can be re-written as:
\begin{align}
    x*_{\mathcal{G}}y=U g(\Lambda_{\theta}) U^T x \label{e3}
\end{align}
where $g(\cdot)$ is the filter function to graph signal $x$, and $\Lambda_{\theta}$ is the diagonal matrix with learnable parameters. \par 
Similar to graph Fourier transform, the graph wavelet transform employs wavelets as spectral basis, the set of wavelets is defined as $\Psi_s = (\psi_{s,1}, \psi_{s,2}, ..., \psi_{s,n})$, where each $\psi_{s,i}$ depends on a scaling parameter $s$ and location $i$. The filter function can then be decomposed in terms of the set of $\Psi_s$:
\begin{align}
    \Psi_s = U \mathbf{G_s} U^T \label{e4}
\end{align}
where $\mathbf{G_s}=diag(g(s\lambda_1), ...,g(s\lambda_n))$ is the filter function with diagonal kernel matrix. $g(s\lambda_i)=e^{-s\lambda_i}$ if using the heat kernel in filter function. \par
The graph wavelet transform can be defined as $\hat{x} = \Psi^{-1}_{s}x$ using the graph wavelets as basis and the inverse graph wavelet transform is $x = \Psi_s \hat{x}$. It is good to note that the inverse of graph wavelets $\Psi^{-1}$ can be calculated by changing the heat kernel $g(s\lambda_i)$ with $g(-s\lambda_i)$. Analogous to Eq.\ref{e2} of the graph Fourier transform, the convolution using graph wavelet transform is written as
\begin{align}
    x*_{\mathcal{G}}y = \Psi_s ((\Psi_s^{-1} y)\odot (\Psi_s^{-1} x)) \label{e5}
\end{align}

Representing $\psi^{-1} y$ with a diagonal kernel matrix $g(\Lambda_{\theta})$,  Eq.\ref{e5} can be represented as:
\begin{align}
    x *_{\mathcal{G}} y = \Psi_s g(\Lambda_{\theta}) \Psi_s^{-1} x \label{e6}
\end{align}

Given the above mathematical details of the similarities and differences between the graph Fourier transform and the graph wavelet transform, we will further compare the graph convolutions with graph Fourier transform and graph wavelet transform in terms of aggregating spatial information.\par
To capture the spatial dependencies of traffic states from the transportation network, it is favorable for the receptive fields of the graph convolution operators to cover the appropriate neighborhood information from the vertex domain. For graph Fourier-based convolution, three commonly used methods are available: 1) Adjacency matrix-based aggregation (\citealp{yu2020forecasting}), 2) Multi-hop neighborhood-based aggregation (\citealp{cui2019traffic}), and 3) Random walk-based aggregation (\citealp{li2017diffusion}). \par 

The adjacency matrix-based method aggregates "1-hop" neighborhood information, sampling information from the first directed connected nodes set ${N}_i$ of the central node $i$. Compared to the adjacency matrix-based method, the multi-hop neighborhood-based method is able to evaluate higher order of proximity by increasing the number of hops. In other words, the multi-hop neighborhood-based method is able to cover a larger range of neighborhood information. Similarly, the random walk-based method is able to handle a certain range of neighborhood information by leveraging finite steps of length over the graph $\mathcal{G}$ from the central node $i$ by sampling the transition probabilities. However, all of these methods assume the constant magnitude of locality for each node in the graph, i.e., the number of hops or finite steps of length, and therefore lack the flexibility to extract spatial information from the neighborhood. \par
\begin{figure}[htp]
    \centering
    \includegraphics[width=0.9\textwidth]{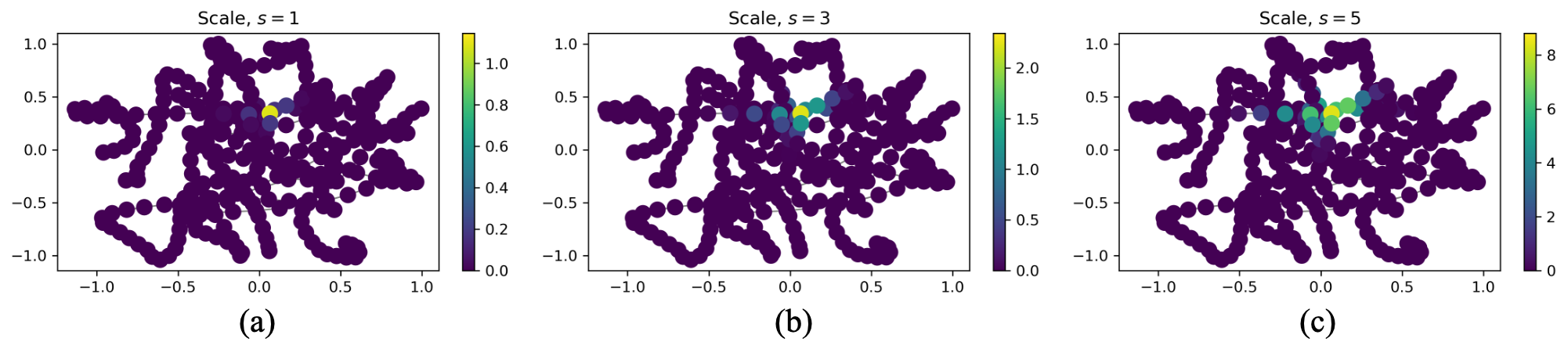}
    \caption{Interaction between the primary node and other nodes at (a) small scale, (b) medium scale and (c) high scale. Value of node represents the interaction between the node and its neighboring node, the larger the value is, the higher the interaction between the nodes.}
    \label{fig:scalesofgraphwavelet}
\end{figure}

Compared to these graph Fourier-based aggregation methods, the localized graph convolution via the graph wavelet transform describes a more flexible neighboring locality. \citealp{hammond2011wavelets} proved that graph wavelets preserve good localization property in the finite scaling parameters. Specifically, the small scales favor well-localized neighborhoods (short diffusion range) and the large scales encode features of coarser neighborhoods (long diffusion range). Fig.\ref{fig:scalesofgraphwavelet} shows the graph wavelets $\psi_{s}$ of each node at different scales in a graph converted by an example transportation network; it is important to note that the layout of the graph is randomly generated based on the topology of road segments, without the use of location information. As mentioned in (\citealp{xu2019graph}), the graph wavelet can be used to denote the different ranges of feature diffusion using different scaling parameter $s$. The range of feature diffusion is represented by the interaction between the primary node and its neighboring nodes. It is clear that as the scaling parameter increases, the feature diffusion range also increases. \par

Beyond the improvement in terms of computational efficiency and the extraction of well-localized features, the graph wavelet transform also has other benefits when used in the problem of traffic prediction (\citealp{xu2019graph}):
\begin{itemize}
    \item The wavelet coefficients $\Psi$ and $\Psi^{-1}$ of the transportation network are usually sparse. Therefore, comparing to graph Fourier transform, the high sparseness of graph wavelet transform can also save computational effort.
    % \item The graph wavelets filtered by heat kernel is localized in the vertex domain on the graph $\mathcal{G}$.
    \item The graph wavelet transform is able to represent the feature diffusion process in a continuous manner. Specifically, the size of a node's neighborhood can be constrained by adjusting the scaling parameter $s$ (Fig.\ref{fig:scalesofgraphwavelet}). A larger value of $s$ corresponds to a larger neighborhood. This feature works even better in solving the problem of traffic prediction where the propagation process of traffic varies in both spatial and temporal dimensions. It provides flexibility in mimicking the real-world traffic propagation process by adjusting the scaling parameter. 
\end{itemize}

However, a single value of scaling parameter still confines a finite locality (diffusion range) of neighborhoods for each node in the graph $\mathcal{G}$. Therefore, we introduce the MS-spatial block to aggregate different levels of spatial features via the graph wavelets at multiple scales. \par

\subsection{Multi-scale spatial block}\label{ssc:ms-spatialblock}
The multi-scale spatial block is formed by stacking multiple graph wavelet nets (GWNs) at different scales. An individual GWN extracts the localized features in a neighborhood at a specific range.  As shown in Eq.\ref{e6}, the construction of GWN can be represented as $\Psi_s g(\Lambda_{\theta}) \Psi_s^{-1} X$. By aggregating multiple (e.g., $e$) GWNs, the architecture of multi-scale spatial block can be written as:
\begin{align}
    H = \sigma(AGG(\Psi_{s_1}\Gamma_{s_1}\Psi^{-1}_{s_1}X, \Psi_{s_2}\Gamma_{s_2}\Psi^{-1}_{s_2}X, ..., \Psi_{s_e}\Gamma_{s_e}\Psi^{-1}_{s_e}X)) \label{e7}
\end{align}
where $H$ denotes the hidden feature of graph convolution layer, $X$ denotes the input feature, $\Gamma_{s}$ denotes the learnable diagonal weight matrix with scaling parameter $s$ and $\psi_{s}$ denotes the graph wavelet coefficient using scaling parameter $s$ with a range in $\{1,e\}$, $AGG(\cdot)$ denotes the aggregation function and $\sigma$ denotes the activation function. To be more specific, the MS-spatial block extracts a short range of features from the neighborhood with small scaling parameters and large range of neighborhood information with large scaling parameters. Compared to graph convolution using single scaling parameter, the MS-spatial block could better tackle different levels of feature information by combining multiple scaling parameters explicitly from the spectral domain. \par 

Accompanied the MS-spatial block with the gated temporal block, the MSGWTCN is able to aggregate the extracted multi-level spatial information at each temporal level. Furthermore, by stacking the gated temporal block and the MS-spatial block in multiple layers, MSGWTCN learns the diffusion of graph signals in multiple spatial-temporal levels by aggregating and combining different scales of spatial and temporal features. \par  

% By stacking multiple graph wavelet and temporal convolution layers, the proposed model can extract both spatial and temporal dependencies at different levels. Specifically, the graph wavelet layer (GWL) at the bottom learns the short-term temporal dependencies while the graph convolution at the top layer extracts long-term temporal information from the gated TCN. Furthermore, by aggregating multiple GWL, MSGWTCN is able to obtain different levels of spatial dependencies. Specifically, GWL extracts the short range of locality of neighborhood using small values of scaling parameters, it captures the global information of neighborhoods with large values of scaling parameters. MSGWTCN then aggregates both extracted short- and long-range neighborhood information at each temporal layer to ensure both local and global spatial information will be learned. To sum up, the proposed multi-scale graph wavelet-based gated temporal convolution network learns the diffusion of graph signals on multiple levels by aggregating and combining different scales of spatial and temporal information.\par
Given the input sequence as $X = (X_{t_1}, X_{t_2}, ..., X_{t_P}) \in \mathbb{R}^{P\times N}$, the prediction in the next step is denoted as $\hat{X}_{t_{P+1}}$. The loss function is defined as follows:
\begin{align}
    MSELoss = MSELoss(\hat{X}_{t_{P+1}}-X_{t_{P+1}}) \label{e10}
\end{align}
where $MSELoss(\cdot)$ is the loss function using mean square error.\par

\section{Numerical Experiments}\label{sc:experiment}
\subsection{Data Description}
\begin{figure}[!h]
  \centering
  \subfloat[Seattle (Source:\citealp{cui2019traffic})]{\includegraphics[width=0.48\textwidth]{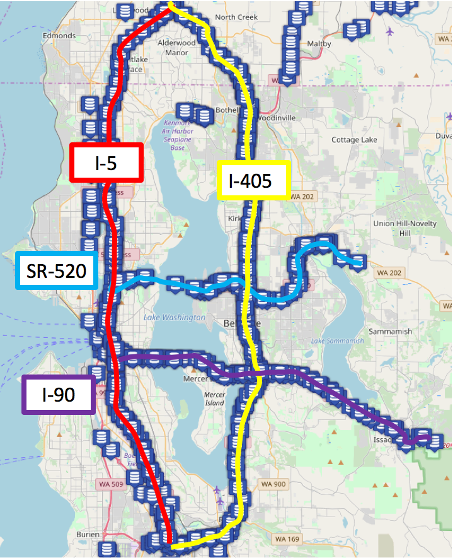}\label{fig:seattlenet}}
  \hfill
  \subfloat[New York City]{\includegraphics[width=0.5\textwidth]{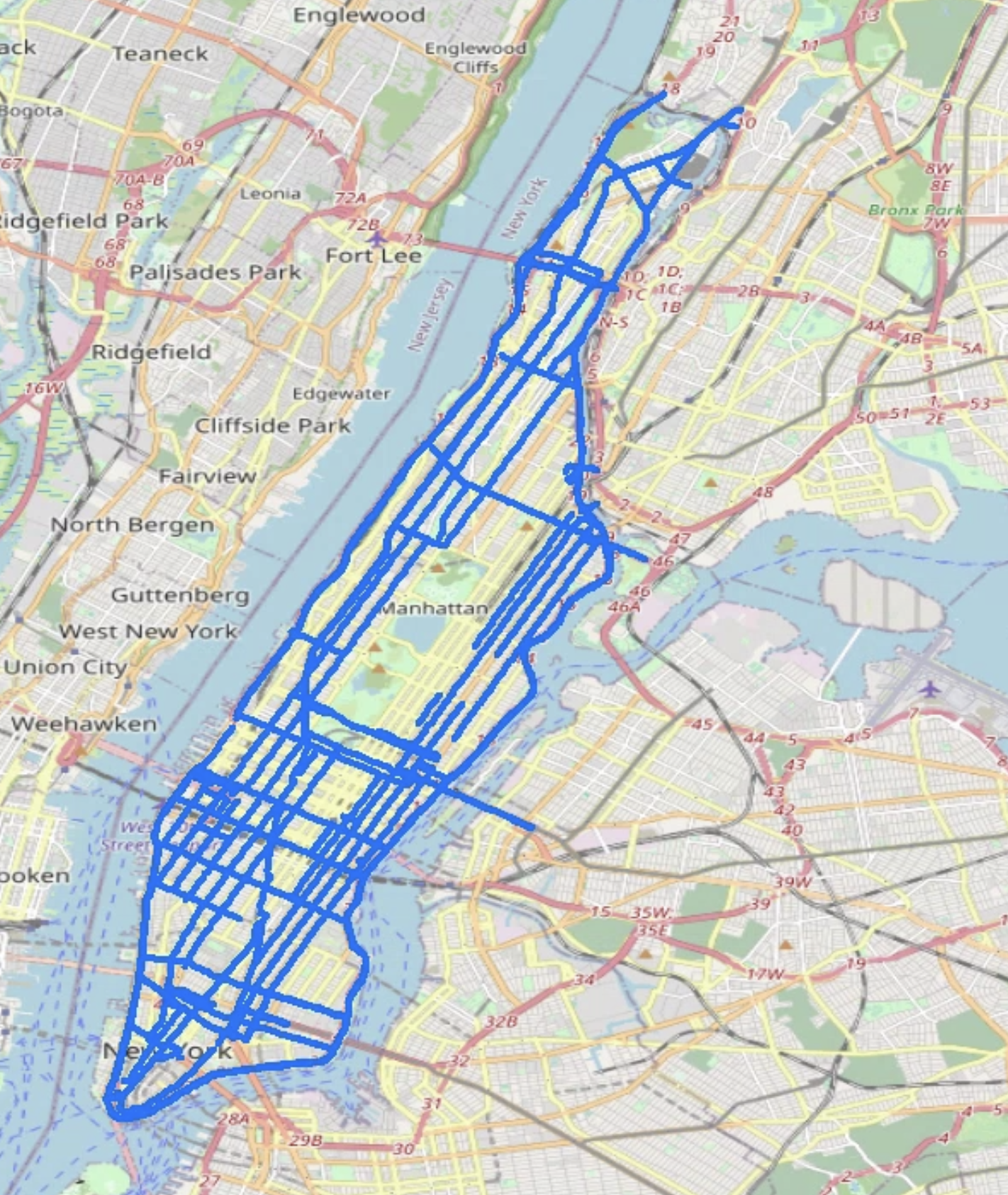}\label{fig:nycnet}}
  \caption{Network layout of Seattle  and New York City.}
  \label{fig:networklayout}
\end{figure}

In this paper, we tested our proposed model on two different transportation networks from two US cities namely, Seattle and New York City. The layout of both networks are shown in Fig.\ref{fig:networklayout}.
\paragraph{Seattle Dataset:} This dataset is cleaned and processed by \citealp{cui2019traffic}. It contains traffic information collected by 323 loop detectors on major highway systems in Seattle. The temporal dimension of this dataset is from January 1st, 2015 to December 31st, 2015 and all the traffic information is represented in 5-minute intervals. 
\paragraph{New York City Dataset:} This dataset is obtained from the National Performance Management Research Data Set (NPMRDS), which includes the traffic speed information collected by probe vehicles. The temporal dimension of this dataset is from January 1st, 2019 to December 31st, 2019. This dataset describes a large scale transportation network in Manhattan, New York City, which contains 1121 roadway segments. The transportation network in this dataset differs from the Seattle dataset in that it contains various types of roadway segments, such as local urban streets, arterials, ramps, bridges, tunnels, highways and so on. Due to the nature of Manhattan's topology, this network is very dense and complex. Similar to Seattle dataset, the speed information is presented in 5-minute intervals. \par 
It should be noted that traffic speed information in both datasets are processed and normalized into the scale of $[0,1]$ in both training and testing phases. 

% \begin{figure}[!tbp]
%   \centering
%   \subfloat[Seattle (Source:(\citealp{cui2019traffic}))]{\includegraphics[width=0.62\textwidth]{Pics/seattle network.png}\label{fig:seattlenet}}
%   \hfill
%   \subfloat[New York City]{\includegraphics[width=0.376\textwidth]{Pics/manhattan network.png}\label{fig:nycnet}}
%   \caption{Network layout of Seattle  and New York City.}
%   \label{fig:networklayout}
% \end{figure}

\subsection{Experimental Settings}
Our proposed model is compared with the following baseline models. The baseline models are mainly based on two prevailing temporal structures: a RNN and a TCN. For RNN models, we selected LSTM, which is a special form of RNNs and other spatial-temporal models in the combination with LSTM (e.g., SGC-LSTM, TGC-LSTM and GWGR). For TCN models, we selected two models that combined the TCN with spatial modules (e.g., STGCN and Graph WaveNet) to predict traffic states. The details for each model are specified as follows:
\begin{itemize}
    \item \textbf{LSTM:}(\citealp{ma2015long}) A long-short term memory neural network.
    \item \textbf{Wavenet:}(\citealp{oord2016wavenet}) A special type of TCN which includes the gating units.
    \item \textbf{SGC-LSTM:}(\citealp{kipf2016semi}) A spectral-based graph convolution neural network adopted from \citealp{kipf2016semi} and combined with LSTM.
    \item \textbf{TGC-LSTM:}(\citealp{cui2019traffic}) A traffic graph convolution neural network with the stack of LSTM and incorporated free-flow reachability matrix. This model adopted k-hop aggregation method to aggregate neighborhood information, the number of hop is set as 3.
    \item \textbf{STGCN:}(\citealp{yu2017spatio}) Spatial-Temporal Graph Convolution Networks, we replicated the model using their proposed source code and kept the same proposed parameters in their model. 
    \item \textbf{GraphWaveNet:}(\citealp{wu2019graph}) An evolved version of Wavenet, which combines Wavenet with spatial modules using graph convolutions via different types of adjacency matrices. We replicated their model using their proposed source code and kept the same proposed parameters. We adopted fixed adjacency matrix and adaptive adjacency matrix and selected whichever the best reported accuracy results.
    \item \textbf{GWGR:}(\citealp{cui2020learning}) Graph Wavelet Gated Recurrent Neural Network, a graph wavelet-based graph convolution networks combining with LSTM as the RNN model. Since there is no open-source code, we replicated the model based on their proposed model structure and reported hyper-parameters. 
\end{itemize}

All the above baseline models are replicated and implemented using Pytorch 1.5.1 and Python 3.6.3. Both baseline models and our proposed models are experimented and trained on a computer with one Intel(R) Xeon(R) E5-2609 CPU @ 2.40GHz and one NVIDIA Titan Xp GPU card and 32GB random-access memory (RAM).\par 

The hyperparameters chosen for MSGWTCN include two parts: parameters for the gated TCN and parameters for GWNs. For the gated TCN, we set kernel size as $(1,1)$ for temporal convolution, a total of 8 TCN layers with dilation factor sequenced and repeated as $[1,2]$, dropout probability as 0.5. For GWNs, we set and test a range of scaling parameters from 0.001 to 6.
During the training phase, we adopted the RMSprop as optimizers with the learning rate as 0.001 and set all the rest of parameters with default values. Moreover, we access the model performances using evaluation metrics including the mean absolute error (MAE) and root mean square error (RMSE), which are calculated as follows:

\begin{align}
    MAE = \frac{1}{n} \sum_{n=1}^{n}  |Y_t- \hat{Y_t}|
\end{align}

\begin{align}
    RMSE = \sqrt{\frac{1}{n} \sum_{n=1}^n (Y_t- \hat{Y_t})^2}
\end{align}

% \begin{align}
%     MAPE = \frac{1}{n} \sum_{n=1}^n |\frac{y_i-\hat{y_i}}{y_i}|
% \end{align}

where $Y_t$ denotes the observed values, $\hat{Y_t}$ denotes the predicted value and $n$ denotes the total number of observations.

\subsection{Experimental Results}

After the test with different scaling parameters ($s\in [0.001,6]$) and numbers (e.g., 1,2,3) of GWNs, we found that the MSGWTCN with the stack of 3 GWNs ($s = 0.85, 3.85, 5.85$) reported the best prediction results as compared to other settings. As shown in Table \ref{predperformance}, the MSGWTCN outperformed all other baseline models in terms of the prediction accuracy. We can observe that for models with the gating recurrent structure, the LSTM reports the worst accuracy with MAE as 3.31 mile/hour in Seattle and 3.87 mile/hour in NYC, while other gating recurrent models report better MAE results. This shows that the incorporation of spatial modules (e.g., graph convolution) can help improve prediction performance. Among these temporal-spatial models with a RNN structure (e.g., SGC+LSTM, TGC+LSTM and GWGR), the GWGR reports the best MAE result as 2.48 mile/hour in Seattle and 2.86 mile/hour in NYC. This may be due to the graph wavelet structure employed by the GWGR can better capture the spatial dependencies in comparison with the rest of graph convolution models (e.g., SGC+LSTM and TGC+LSTM). \par 
For models with the TCN structures, the GraphWavenet generates better MAE as 2.39 mile/hour in Seattle and 2.66 mile/hour in NYC while the Wavenet showed a defective result of MAE as 2.53 mile/hour in Seattle and 2.72 mile/hour in NYC. The use of the adaptive adjacency matrix in the GraphWavenet helped capture the dynamics of the spatial dependencies in the test networks. However, the MSGWTCN outperforms both TCN-based models (e.g., Wavenet and GraphWavenet) and gives the best MAE and RMSE for both the Seattle and NYC datasets. \par 
Comparing temporal structures, (e.g., TCN and RNN), the TCN is observed to have a better accuracy than RNN models, both in the temporal-only models and spatial-temporal models. The reason may be that the TCN could employ a more efficient storage of the temporal information by dilating different steps of the sequential data to capture both short- and long-term temporal dependencies. 
Comparing spatial-temporal models that employed the graph wavelet to deal with the spatial dependencies, the MSGWTCN has better prediction performance than GWGR for both the Seattle and NYC datasets. One possible reason is due to the power of multi-scale graph wavelet layers that enabled MSGWTCN to capture the spatial dependencies in multiple diffusion levels (e.g., local, intermediate and global). Moreover, the design of the dilation in the TCN also helps the trained model to better capture multiple levels of the temporal dependencies through the convolution of multiple sizes of the sequence data (e.g., short-, intermediate and long-term).
 
\begin{table}[!ht]
    \centering
    \caption{Comparison of prediction performance using other baseline models}
    \label{predperformance}
    \small
    \begin{tabular}{|c|c|c|c|c|}
    \hline
    \multirow{2}{*}{Model} & \multicolumn{2}{c|}{Seattle Dataset} &
    \multicolumn{2}{c|}{NYC Dataset}\\
    \cline{2-5}
    & MAE & RMSE & MAE & RMSE \\
    \hline
    LSTM & 3.31 & 0.40 & 3.87 & 0.54\\
    \hline
    Wavenet & 2.53 & 0.30 & 2.72 & 0.45\\
    \hline
    SGC-LSTM & 2.64 & 0.38 & 3.80 & 0.51\\
    \hline
    STGCN & 2.64 & 0.35 & 3.33 & 0.49\\
    \hline
    TGC-LSTM & 2.57 & 0.33 & 3.09 & 0.46\\
    \hline
    GraphWavenet & 2.39 & 0.28 & 2.66 & 0.44\\
    \hline
    GWGR & 2.48 & 0.30 & 2.86 & 0.46\\
    \hline
    % GWTCN & 2.32 & 0.27 & 2.61 & 0.43\\
    % \hline
    \textbf{MSGWTCN} & \textbf{2.28} & \textbf{0.27} & \textbf{2.59} & \textbf{0.43}\\
    \hline
    \end{tabular}
\end{table}

\subsection{Computational Complexity}
As Eq.\ref{e7} shows that the weight parameter $\Gamma_s$ is formed as a weight matrix with only diagonal values available to learn, which reduces the total number of weight parameters to be learned in the training process. Moreover, we employed the Chebyshev polynomial method (\citealp{hammond2011wavelets}) to efficiently approximate the graph wavelets (e.g., $\Psi$ and $\Psi^{-1}$). In this work, we adopted the approximation order of the Chebyshev polynomials $k$ as 3. In this way, we saved the effort of calculating the decomposition of the eigenvalues and eigenvectors of the graph Laplacian $\mathbf{L}$. Furthermore, the sparsity of the graph wavelets reduces the computation complexity during the training process.

\section{Result Analysis}\label{sc:result}
In this section we investigate the scalability of our designed MSGWTCN structure for the prediction of traffic states by analyzing the MS-spatial block and gated TCN. To verify the feasibility of the MSGWTCN, we firstly need to prove that the MSGWTCN is sensitive to different values of the scaling parameters. Specifically, we will test the prediction performance of the MSGWTCN in the different scaling parameters to show its sensitivity in the feature learning through MAE. Second, we investigate the effect of the feature extraction for the MS-spatial block by comparing the prediction performances through different settings in GWNs. Third, we perform two kinds of weight analysis to show that the combination of the MS-spatial block and gated TCN are feasible to the multi-scale traffic feature extraction. At last, we conduct an ablation experiment to explore the sensitivity of the MSGWTCN to transportation networks with different road type configurations.

\subsection{Sensitivity analysis of scaling parameters}
\begin{figure}[htp]
    \centering
    \includegraphics[width=0.7\textwidth]{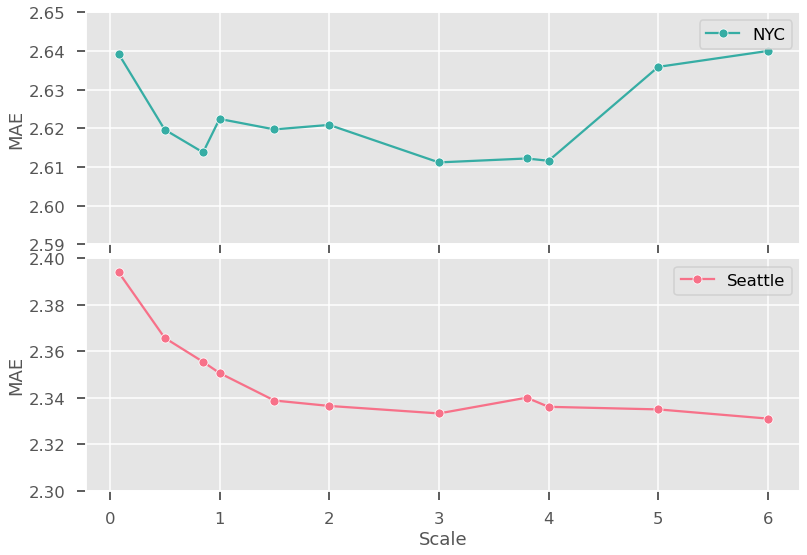}
    \caption{Prediction performance (MAE) to different values of scaling parameter}
    \label{fig:sensitivity}
\end{figure}

A sensitivity analysis is conducted to see whether the prediction performance of the MSGWTCN varies with different values of scales when employing a single GWN. As shown in Fig.\ref{fig:sensitivity}, the MSGWTCN is sensitive to different values of the scaling parameters in terms of MAE and has a different trend for the NYC and Seattle datasets. For Seattle, the MSGWTCN shows a monotonous trend wherein the MAE performance becomes better as the value of scaling parameter increases (MAE drops from 2.4 to 2.33). The potential rationale behind this may be that the Seattle dataset mainly consists of the highway network, and is therefore significant to a larger spatial receptive field (e.g., neighborhood aggregation) rather than a smaller one. \par 

In contrast, for the NYC dataset, the MAE decreases firstly as the scaling parameter increases and then starts increasing as the scaling parameter increases after $s=3$. One potential reason could be that the NYC dataset consists of a very complicated transportation network of not only highways but also other types of road segments such as local urban streets, ramps, tunnel entrances, bridges, and so on. Therefore, the use of single GWN with single scale may not efficiently extract the most appropriate receptive field in capturing the spatial dependencies.\par 

This sensitivity analysis also provides a recommendation for the selection of scaling parameters in extracting the spatial features. For example, for the Seattle dataset, $s \in [1.5,5]$ may work better by providing good MAE performances and $s \in [0.85, 4]$ may show better results for the NYC dataset.

\subsection{The comparison of stacking multi-scale graph wavelets}
We will discuss the effect of the fused feature learning by stacking multiple GWNs through the MSGWTCN.
Table \ref{multi-scale_mae} shows the MAE results of MSGWTCNs with different scaling parameter $s$. The number "3" in the MSGWTCN-3 refers to the fact that the MSGWTCN employs three GWNs to learn the multi-scale spatial information. Similarly, the MSGWTCN-1 means that the MSGWTCN utilizes single GWN and the MSGWTCN-2 means 2 GWNs are used in the feature learning process. The specific settings of scaling parameter $s$ is given in brackets. Overall, it is easy to observe from Table \ref{multi-scale_mae} that the model prediction performance becomes better as we aggregate features at more scales. To be specific, the MSGWTCN-1 were tested with three different scales and the MAEs in both cities are higher than the MSGWTCN-2, which stacked two GWNs ($s=0.85, 3.85$). Furthermore, after stacking three GWNs with three different scales ($s=0.85, 3.85, 5.85$), the MSGWTCN-3 achieved the best prediction results as compared to both MSGWTCN-1 and MSGWTCN-2 in the two cities, 2.28 mile/hour for Seattle and 2.59 mile/hour for NYC, respectively. \par 
It is easy to confirm that extracting multi-scale neighborhood information is important when dealing with the complex transportation networks and helps improve the prediction of traffic states for such networks. 

\begin{table}[!ht]
    \centering
    \caption{MAE results of stacking multi-scale GWNs}
    \label{multi-scale_mae}
    \small
    \begin{tabular}{|c|c|c|}
    \hline
    Model & Seattle & NYC\\
    \hline
    MSGWTCN-1 (s=0.85) & 2.35 & 2.61 \\
    \hline
    MSGWTCN-1 (s=3.85) & 2.34 & 2.61 \\
    \hline
    MSGWTCN-1 (s=5.85) & 2.33 & 2.64\\
    \hline
    MSGWTCN-2 (s=0.85, 3.85) & 2.32 & 2.60\\
    \hline
    % MSGWTCN-3 (s=0.08, 0.58, 1.58) & 2.32 & TBA\\
    % \hline
    % MSGWTCN-3 (s=1, 2, 3) & 2.33 & TBA\\
    % \hline
    % MSGWTCN-3 (s=4, 4.5, 5) & 2.31 & TBA\\
    % \hline
    % MSGWTCN-3 (s=0.85, 0.85, 0.85) & 2.34 & 2.62\\
    % \hline
    % MSGWTCN-3 (s=3.85, 3.85, 3.85) & 2.32 & 2.61\\
    % \hline
    % MSGWTCN-3 (s=5.85, 5.85, 5.85) & 2.32 & 2.60\\
    % \hline
    \textbf{MSGWTCN-3 (s=0.85, 3.85, 5.85)} & \textbf{2.28} & \textbf{2.59}\\
    \hline
    \end{tabular}
\end{table}

\subsection{Multi-scale graph weight analysis}
In this subsection, we conduct further investigation into the effect of multi-scale feature learning by extracting and analyzing the learned weight parameters from the MSGWTCN-3. We will use the term "TCN" for abbreviation of the gated TCN in the following discussion.\par 
% With the power of stacked gated TCN (we will use the term "TCN" for abbreviation in the following discussion) layers, MSGWTCN is able to extract feasible receptive fields in different temporal levels. As mentioned in Section.\ref{ssc:gtcn}, the stacked TCN can learn both short- and long-term temporal correlations at different layers. Therefore, by extracting the learned weight parameters from each layer, we are able to investigate the effect of multi-scale GWNs at different temporal levels.\par 
To be specific, the weight parameters of the MSGWTCN-3 include three diagonal weight matrices ($\Gamma_{s_1}^i, \Gamma_{s_2}^i, \Gamma_{s_3}^i$) for $i_{th}$ TCN layer ($i \in [1,8]$). We then convert the learned diagonal weight matrices to the graph wavelet weight matrices by multiplying graph wavelet matrices (e.g., $\Psi_s$ and $\Psi_s^{-1}$) of small, medium and large values of scaling parameters. The graph wavelet weight matrices are represented as $\Psi_{s_1}\Gamma_{s_1}^i\Psi_{s_1}^{-1}, \Psi_{s_2}\Gamma_{s_2}^i\Psi_{s_2}^{-1}$ and $\Psi_{s_3}\Gamma_{s_3}^i\Psi_{s_3}^{-1}$. \par
\begin{figure}[!tbp]
  \centering
  \subfloat[TCN Layer 1-4.]{\includegraphics[width=0.5\textwidth]{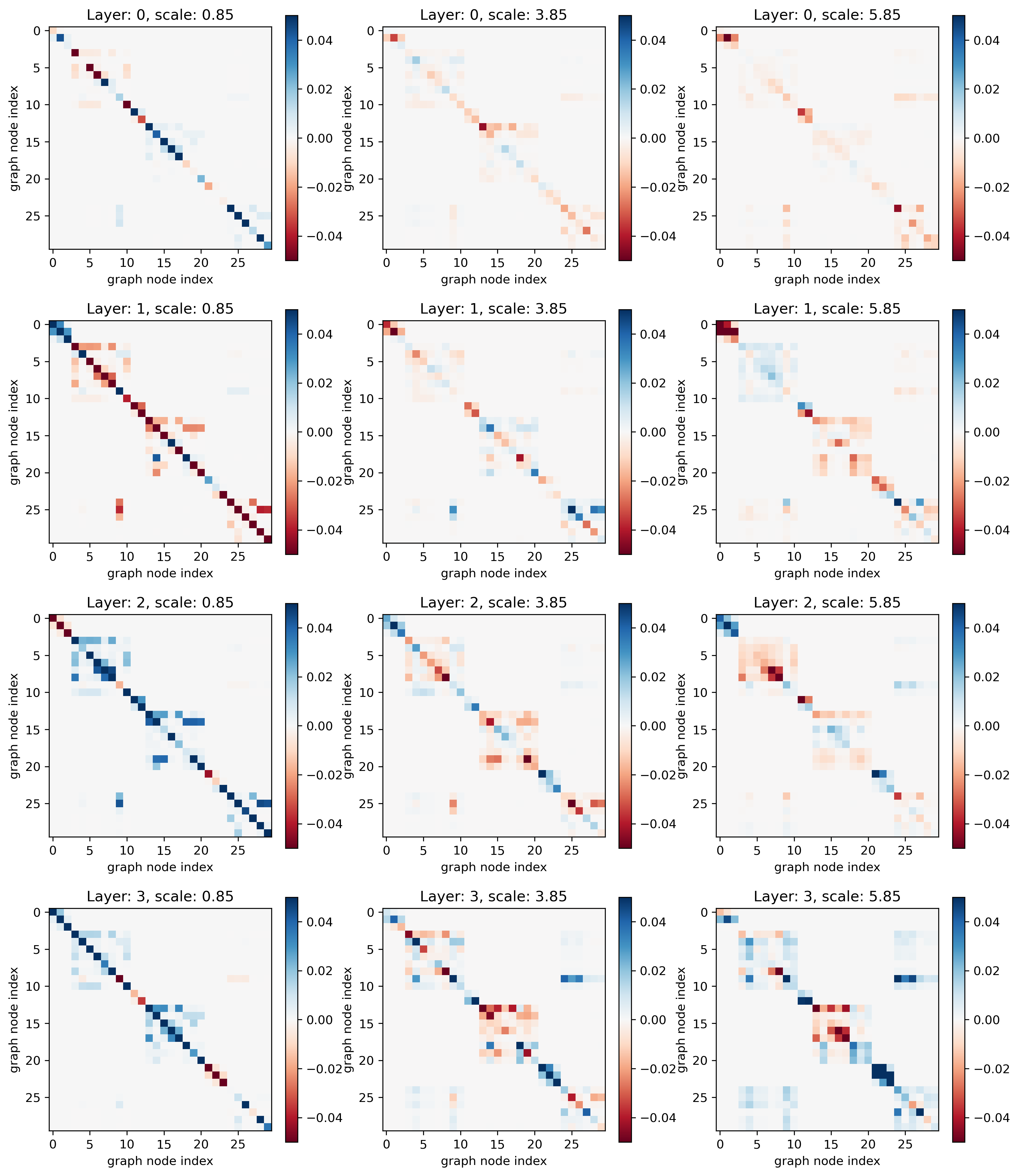}\label{fig:dilation1}}
  \hfill
  \subfloat[TCN Layer 5-8.]{\includegraphics[width=0.5\textwidth]{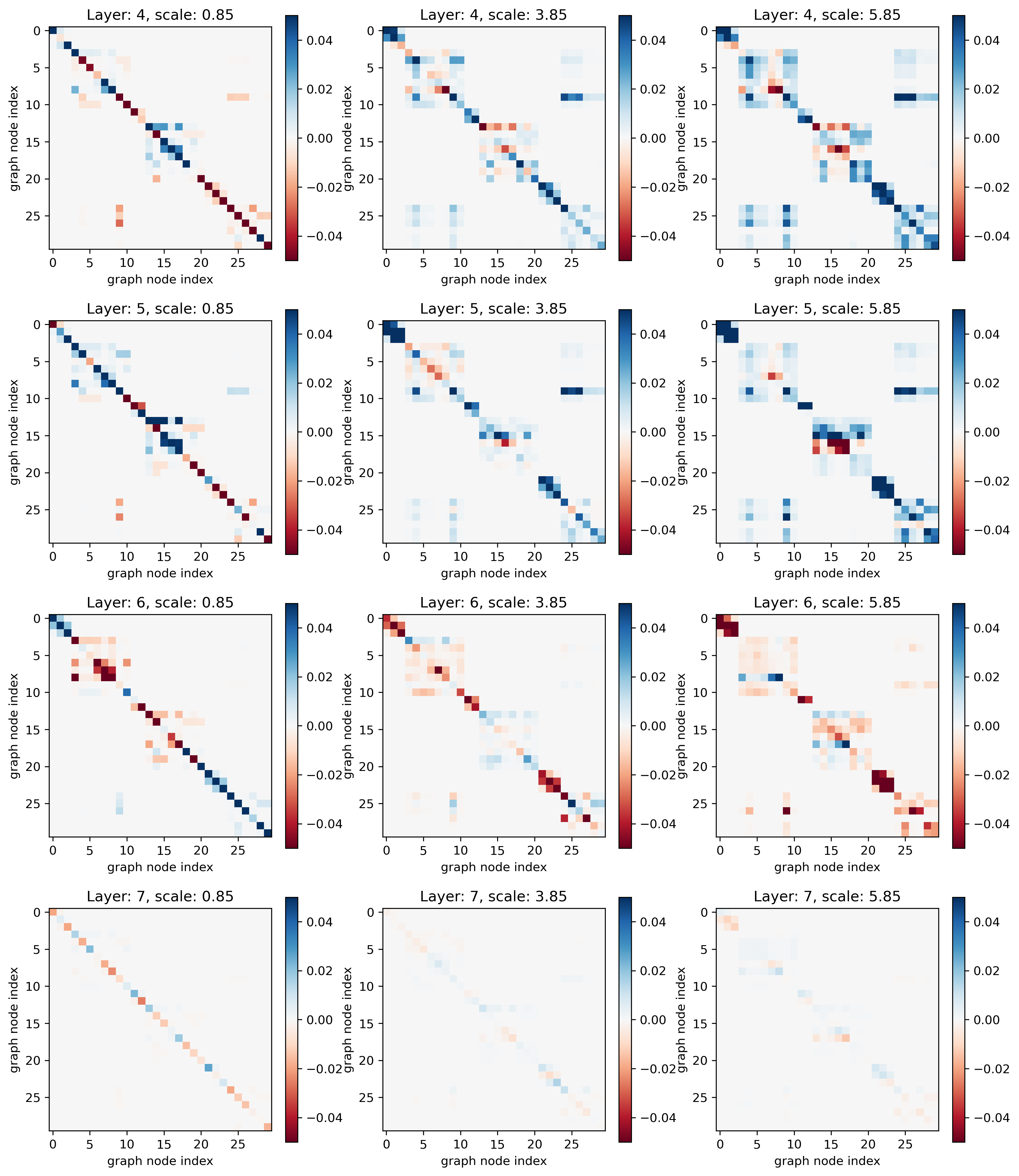}\label{fig:dilation2}}
  \caption{First 30 nodes of graph wavelet weight matrices at TCN layers. Red and blue represent the values of the graph wavelet weight matrices, the darker color represents the higher value and vice versa.}
  \label{fig:WeightMat_Temporal}
\end{figure}

 
We selected the NYC dataset as an example since it consists of a complex transportation network containing various types of the road segments. For demonstration purposes, we extracted the first 30 graph nodes and show the graph wavelet weight matrices in Fig.\ref{fig:WeightMat_Temporal}. Each graph wavelet weight matrix contains 30 rows and 30 columns, each row-wise picture presents the weight matrices at each TCN layer, each column presents each converted graph wavelet weight matrices as scale $s$ ($s \in (0.85, 3.85, 5.85)$). \par 

The absolute values of elements in the graph wavelet weight matrices indicate the contribution from the graph node to the prediction of future traffic states. The higher the absolute value is, the more contribution made from the graph node to the prediction. The diagonal values of the graph wavelet weight matrices represent the contribution from the graph node to its future states. The non-diagonal values indicate the interactions between the graph node and its neighboring nodes. Specifically, the larger the area with the obvious red or blue colors, the more graph nodes are observed with the spatial interactions. For example, it is easy to observe that for each layer, the small scale ($s=0.85$) graph wavelet matrices showed the best locality of the spatial interaction since the obvious colors of the heat map mostly focused on diagonal dots. For the medium scale ($s=3.85$) graph wavelet weight matrices, the colors of the diagonal values are not as obvious as the small scale but the clusters of colors are broader. The graph wavelet weight matrices with high scale showed the lightest color of the diagonal dots but the color cluster is the broadest. \par 

\begin{figure}[h]
    \centering
    \includegraphics[width=0.8\textwidth]{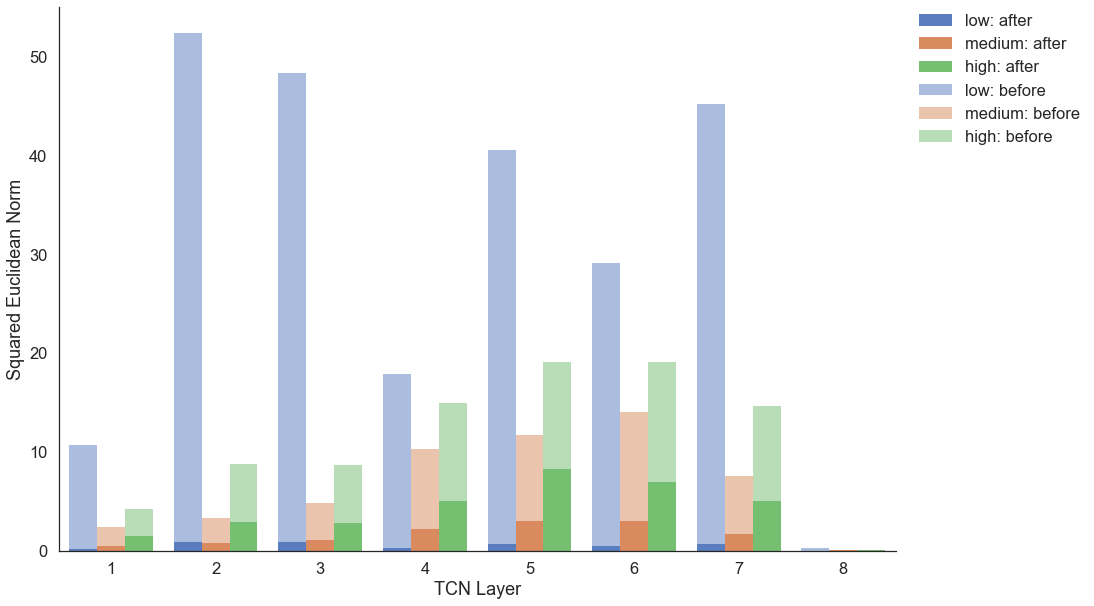}
    \caption{Squared Euclidean norm of graph wavelet matrices at TCN layers. The blue bar presents the squared norms of low scale ($s=0.85$), orange presents the medium scale ($s=3.85$) and green refers to the high scale ($s=5.85$). The light color is used to indicate the squared norms before forcing diagonal values to 0, and the dark color indicates the squared norms after forcing diagonal values to 0.}
    \label{fig:matrixl2norm}
\end{figure}

To further confirm this finding, we conduct a before-after experiment by forcing the diagonal values of the graph wavelet weight matrices to 0. We calculate the squared Euclidean norm of the graph wavelet weight matrix,in order to measure the contribution of a weight parameter in generating the model outputs. The larger norm indicates more contribution of a weight parameter and vice versa. The calculated squared norms are shown in Fig.\ref{fig:matrixl2norm}.\par 

Before forcing the diagonal values to 0, the low-scale weight matrices showed the largest squared Euclidean norm across all 8 TCN layers. This indicates that the low-scale weight matrices contribute the most in the prediction process. However, after forcing diagonal values to 0, a sharp drop in the squared norm values among the low-scale weight matrices was found. On the contrary, the removal of diagonal values (force to 0) showed the relatively smaller impact to the medium- and high-scale weight matrices and the high scale had the largest squared norms over all TCN layers. This confirms our findings that the low-scale graph wavelet weight matrices contribute more to the graph nodes themselves and thus are good at extracting the locality in spatial dependencies; The medium-scale graph wavelet weight matrices involved more graph nodes and thus captured the intermediate level of the spatial information; The high-scale graph wavelet weight matrices contributed the most to interact with other graph nodes and were affected the least by the node itself. The coherent working of different scales of the graph wavelet weight matrices ensures the MSGWTCN captures the spatial dependencies in multiple scales of the neighborhood information simultaneously. \par

\subsection{Significant link identification at multiple scales}
\begin{figure}[h]%[!tbp]
  \centering
  \subfloat[Low scale.]{\includegraphics[width=0.333\textwidth]{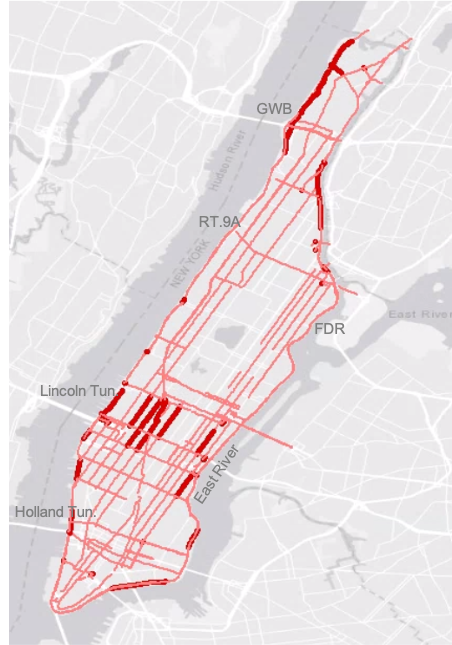}\label{fig:lowscalemanhattan}}
  \hfill
  \subfloat[Medium scale.]{\includegraphics[width=0.329\textwidth]{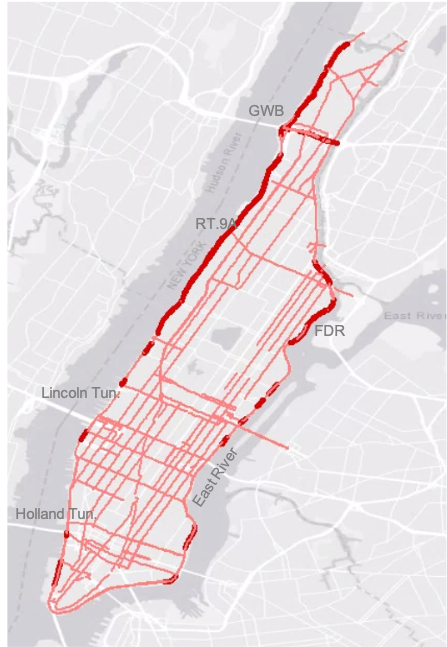}\label{fig:medscalemanhattan}}
  \hfill
  \subfloat[High scale.]{\includegraphics[width=0.33\textwidth]{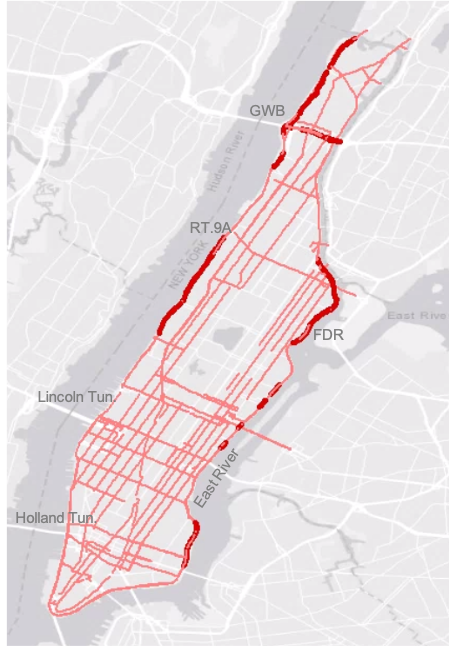}\label{fig:highscalemanhattan}}
  \caption{Most significant road segments in NYC. The highlighted dark red links are significant links}
  \label{fig:multi-scalemanhattan}
\end{figure}

The squared Euclidean norms for each individual road link (e.g., row-wise elements in graph wavelet weight matrix) and summed up across 8 TCN layers in NYC dataset. The top 5 percentage squared norms links were highlighted in dark red on the map in Fig.\ref{fig:multi-scalemanhattan}. Intuitively, the road links with the high squared norms depict the road links that are significantly influential to the transportation network. Most of the most influential road links identified by the low scale graph wavelet weights are urban streets, while the medium scale and high scale graph wavelet weights identified mainly highway links around Manhattan, NYC.\par 
To better interpret the identified road links, we raised up several empirical cases. For example, the low scale graph wavelet weights mainly identified urban streets around midtown and downtown Manhattan, around the areas close to the bridge and tunnel crossings connecting New Jersey and NYC (e.g., Lincoln Tunnel and Holland Tunnel), which are empirically known as the congested areas due to both inbound and outbound traffic. Besides the inter-borough traffic, these urban streets also carry the intra-borough traffic between uptown and downtown Manhattan. As compared to the low scale weights, both medium and high scale weights identified road links of the two major highways in Manhattan (New York State Route 9A and Franklin D. Roosevelt (FDR) drive). Specifically, both medium and high scale weights identified the most influential links that connect the local and regional traffic to George Washington Bridge (GWB), one of the busiest bridges in the US, as well as the FDR highway links near a majority of the east river crossings. \par 
In summary, the multi-scale settings allow for scalability when using the MSGWTCN to identify multiple types of the road links at the same time and help interpret the model results for large transportation networks that contain the complex road structures, such as a mix of dense local streets and highways or networks with limited access.

\subsection{Transportation network ablation experiment}
\begin{figure}[!h]
    \centering
    \includegraphics[width=0.6\textwidth]{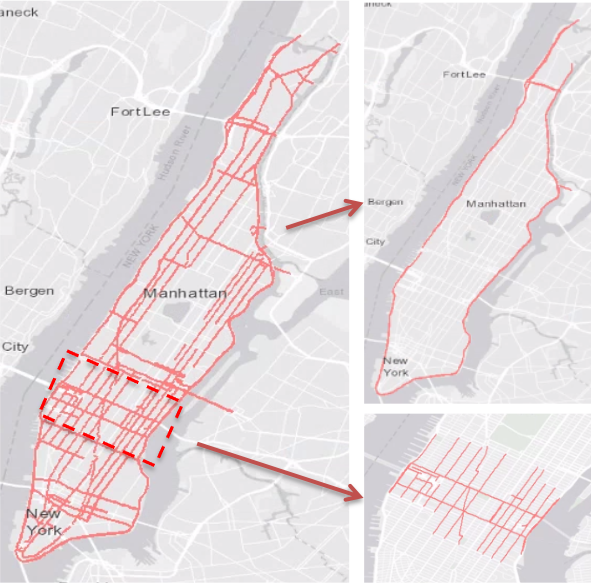}
    \caption{Network ablation experiment for Manhattan}
    \label{fig:networkablation}
\end{figure}

Beyond a discussion of the MSGWTCN at the link level and neighborhood level, we would also like to explore the sensitivity of the MSGWTCN to different types of the transportation networks. To do this, we conduct a network ablation experiment using the NYC dataset. We dissolve partial areas from the whole transportation network in terms of the types of roadway systems to get two types of sub-network: a highway-only network (224 road links) and an urban street network (155 road links) (Fig.\ref{fig:networkablation}). We then experiment the different sets of scales in the MSGWTCN-3, such as three small scales ($s=0.85,0.85,0.85$), three medium scales ($s=3.85,3.85,3.85$), three high scales ($s=5.85,5.85,5.85$) and the group of small, medium and high scales ($s=0.85,3.85,5.85$).  \par

We adopted MAE as the prediction metric to evaluate the different settings of multi-scale GWNs in MSGWTCN-3 ( Table.\ref{ablation_mae}). It is easy to observe the trend in the highway network: MAE decreased as $s$ increased and reached the best MAE at the combination of the small-, medium- and large- scales. This is in line with the results reported by the Seattle dataset, which showed the best MAE at the scale set $s=0.85,3.85,5.85$. Nevertheless, midtown Manhattan showed a different trend. The combination of the small, medium and large scales reported the worst MAE, while the three high scales showed the best MAE.\par 

\begin{table}[!h]
    \centering
    \caption{MAE performance of ablation experiment}
    \label{ablation_mae}
    \small
    \begin{tabular}{|c|c|c|c|}
    \hline
    Model & Seattle & NYC (Highway) & NYC (Midtown)\\
    \hline
    MSGWTCN-3 (s=0.85, 0.85, 0.85) & 2.34 & 3.78 & 2.56\\
    \hline
    MSGWTCN-3 (s=3.85, 3.85, 3.85) & 2.32 & 3.74 & 2.55\\
    \hline
    MSGWTCN-3 (s=5.85, 5.85, 5.85) & 2.31 & 3.74 & \textbf{2.54}\\
    \hline
    MSGWTCN-3 (s=0.85, 3.85, 5.85) & \textbf{2.28} & \textbf{3.73} & 2.59\\
    \hline
    \end{tabular}
\end{table}
% In summary, we find that the multi-scale settings are sensitive to different types of transportation networks. Specifically, networks with mainly a highway system see the best prediction performance with multi-scale GWNs with small, medium and high scaling parameters. However, a transportation network with mainly urban streets does not favor the combination of different scales but three large scales only.\par %Road type configurations in network are sensitive to scale settings.
This result can be due to the aggregation levels of different scales of graph wavelets. As mentioned in Fig.\ref{fig:matrixl2norm}, the small scale of graph wavelets extract well-localized spatial features and thus focus on the states of the road link itself, while the large scale of graph wavelets favor the interactions of the road segments in a larger neighborhood. The traffic states on a highway road segment is usually affected mainly by the segment itself and its direct adjacent neighbor links and is not easily affected by other road segments during the normal traffic conditions.  During the congestion periods (e.g., accident, peak hour), the interaction between the road links becomes strong and one road link is likely affected by some further segments. Settings with the small, medium and high scales can thus better adapt to different traffic conditions by extracting multiple levels of the spatial interactions.\par 
For a dense urban network such as Midtown Manhattan, however, the interaction between road segments is always strong due to the heavy daily traffic. The queue spill-back due to the recurrent congestion and frequent traffic events such as accidents, curb activities, and traffic lights which could also lead to an area-wide interactions among the road segments. Therefore, the accommodation of the high scale set to always focus on a larger neighborhood can better help the urban street network in the prediction rather than using other scale combinations.\par 
% Other than the different road type configurations between these two networks, another potential reason could be the size/scale of the experimented networks. For example, the highway networks of both Seattle and NYC used in this study are large regions, while midtown Manhattan only contains a small road network with 155 road links only. The road links in a small-scale transportation network may interact more with each other and therefore the set of high scales can better extract the spatial features. \par 
% In summary, the multi-scale GWNs allows us to flexibly adapt the scaling parameters to the structure of the transportation network. We show that for the two tested highway networks, multi-scale settings show better prediction performance with combinations of small, medium and high scales. For dense urban street network, multi-scale settings with large scales achieve the best prediction performance.

\section{Conclusion} \label{sc:conclusion}

In this study, we propose a multi-scale graph wavelet based temporal convolution network (MSGWTCN) to predict the traffic speed information in the complex transportation networks. As compared to other methods of aggregating the spatial information from a neighborhood in a fixed/discrete manner, the graph wavelet provides a flexible way to extract the spatial dependencies in a continuous manner. Moreover, we propose a MS-spatial block to aggregate different scales of the graph wavelets at the same time. The MS-spatial block employs multi-scale GWNs to extract the spatial information at different levels. The MSGWTCN outperformed other prevailing deep learning based methods in the prediction performance. Moreover, two kinds of weight analyses were conducted and showed that the multi-scale graph wavelets are able to extract local, intermediate and global levels of the spatial information simultaneously. We also show that the MSGWTCN is able to deal with the large transportation networks that contain the complex road structures, and that multi-scale settings can help the model to identify different types of the road segments. At last, we conduct a network ablation experiment to study the sensitivity of the MSGWTCN to different types of the road networks. As a result, we found out that the dense urban street network showed the best performance with the combination of only high scales, while the large highway network responded better to the combination of small, medium and high scales.  \par

There are still several possibilities to improve our proposed model. In the future, we will study the effectiveness of different graph wavelets and try to make the MSGWTCN to be adaptive to the different network structures in an automated manner. Furthermore, we will investigate the correlation between the graph wavelets and external factors in the presence of the real-world  traffic conditions, such as traffic incidents, and explore if the graph wavelets can help extract the change of the road conditions such as road closures due to the occurrence of traffic accidents.

\section*{CRediT authorship contribution statement}
\textbf{Zilin Bian}: Conceptualization, Data curation, Methodology, Formal analysis, Investigation, Visualization, Writing - original draft, Writing - review \& editing. \textbf{Jingqin Gao}: Formal analysis, Investigation, Writing - original draft, Writing - review \& editing. \textbf{Kaan Ozbay}: Supervision, Writing - review \& editing. \textbf{Zhenning Li}: Writing - review \& editing.

\section*{Acknowledgements}
% This study was partially supported by the University Transportation Research Center (UTRC) at the City University of New York, CIDNY (Coordinated Intelligent Transportation Systems Deployment in New York City) program, and C2SMART, a Tier 1 University Transportation Center at New York University. The contents of this paper only reflect views of the authors who are responsible for the facts and do not represent any official views of any sponsoring organizations or agencies. \par
This work was supported by the Connected Cities for Smart Mobility towards Accessible and Resilient Transportation (C2SMART) Center, a Tier 1 University Center awarded by U.S. Department of Transportation under the University Transportation Centers Program. The contents of this paper reflect the views of the authors, who are responsible for the facts and the accuracy of the information presented herein. This work is funded, partially or entirely, by a grant from the U.S. Department of Transportation’s University Transportation Centers Program. However, the U.S. Government assumes no liability for the contents or use thereof.\par 

% \section*{Appendix} \label{sc:appendix}
% TBD.

%\bibliographystyle{apacite}
%\bibliographystyle{apa}
%\nocite{*}

\bibliography{mybib}

\begin{thebibliography}{41}
\providecommand{\natexlab}[1]{#1}
\providecommand{\url}[1]{\texttt{#1}}
\expandafter\ifx\csname urlstyle\endcsname\relax
  \providecommand{\doi}[1]{doi: #1}\else
  \providecommand{\doi}{doi: \begingroup \urlstyle{rm}\Url}\fi

\bibitem[Bogaerts et~al.(2020)Bogaerts, Masegosa, Angarita-Zapata, Onieva, and
  Hellinckx]{bogaerts2020graph}
T.~Bogaerts, A.~D. Masegosa, J.~S. Angarita-Zapata, E.~Onieva, and
  P.~Hellinckx.
\newblock A graph cnn-lstm neural network for short and long-term traffic
  forecasting based on trajectory data.
\newblock \emph{Transportation Research Part C: Emerging Technologies},
  112:\penalty0 62--77, 2020.

\bibitem[Chang et~al.(2020)Chang, Rong, Xu, Huang, Sojoudi, Huang, and
  Zhu]{chang2020spectral}
H.~Chang, Y.~Rong, T.~Xu, W.~Huang, S.~Sojoudi, J.~Huang, and W.~Zhu.
\newblock Spectral graph attention network.
\newblock \emph{arXiv preprint arXiv:2003.07450}, 2020.

\bibitem[Chen et~al.(2011)Chen, Hu, Meng, and Zhang]{chen2011short}
C.~Chen, J.~Hu, Q.~Meng, and Y.~Zhang.
\newblock Short-time traffic flow prediction with arima-garch model.
\newblock In \emph{2011 IEEE Intelligent Vehicles Symposium (IV)}, pages
  607--612. IEEE, 2011.

\bibitem[Chen et~al.(2019)Chen, Li, Teo, Zou, Wang, Wang, and
  Zeng]{chen2019gated}
C.~Chen, K.~Li, S.~G. Teo, X.~Zou, K.~Wang, J.~Wang, and Z.~Zeng.
\newblock Gated residual recurrent graph neural networks for traffic
  prediction.
\newblock In \emph{Proceedings of the AAAI Conference on Artificial
  Intelligence}, volume~33, pages 485--492, 2019.

\bibitem[Cho et~al.(2014)Cho, Van~Merri{\"e}nboer, Bahdanau, and
  Bengio]{cho2014properties}
K.~Cho, B.~Van~Merri{\"e}nboer, D.~Bahdanau, and Y.~Bengio.
\newblock On the properties of neural machine translation: Encoder-decoder
  approaches.
\newblock \emph{arXiv preprint arXiv:1409.1259}, 2014.

\bibitem[Chu et~al.(2019)Chu, Lam, and Li]{chu2019deep}
K.-F. Chu, A.~Y. Lam, and V.~O. Li.
\newblock Deep multi-scale convolutional lstm network for travel demand and
  origin-destination predictions.
\newblock \emph{IEEE Transactions on Intelligent Transportation Systems},
  21\penalty0 (8):\penalty0 3219--3232, 2019.

\bibitem[Cui et~al.(2019)Cui, Henrickson, Ke, and Wang]{cui2019traffic}
Z.~Cui, K.~Henrickson, R.~Ke, and Y.~Wang.
\newblock Traffic graph convolutional recurrent neural network: A deep learning
  framework for network-scale traffic learning and forecasting.
\newblock \emph{IEEE Transactions on Intelligent Transportation Systems},
  21\penalty0 (11):\penalty0 4883--4894, 2019.

\bibitem[Cui et~al.(2020)Cui, Ke, Pu, Ma, and Wang]{cui2020learning}
Z.~Cui, R.~Ke, Z.~Pu, X.~Ma, and Y.~Wang.
\newblock Learning traffic as a graph: A gated graph wavelet recurrent neural
  network for network-scale traffic prediction.
\newblock \emph{Transportation Research Part C: Emerging Technologies},
  115:\penalty0 102620, 2020.

\bibitem[Dai et~al.(2019)Dai, Fu, Zhao, Zhang, Lin, Wang, and
  Li]{dai2019deeptrend}
X.~Dai, R.~Fu, E.~Zhao, Z.~Zhang, Y.~Lin, F.-Y. Wang, and L.~Li.
\newblock Deeptrend 2.0: A light-weighted multi-scale traffic prediction model
  using detrending.
\newblock \emph{Transportation Research Part C: Emerging Technologies},
  103:\penalty0 142--157, 2019.

\bibitem[Dauphin et~al.(2017)Dauphin, Fan, Auli, and
  Grangier]{dauphin2017language}
Y.~N. Dauphin, A.~Fan, M.~Auli, and D.~Grangier.
\newblock Language modeling with gated convolutional networks.
\newblock In \emph{International conference on machine learning}, pages
  933--941. PMLR, 2017.

\bibitem[Do et~al.(2019)Do, Vu, Vo, Liu, and Phung]{do2019effective}
L.~N. Do, H.~L. Vu, B.~Q. Vo, Z.~Liu, and D.~Phung.
\newblock An effective spatial-temporal attention based neural network for
  traffic flow prediction.
\newblock \emph{Transportation research part C: emerging technologies},
  108:\penalty0 12--28, 2019.

\bibitem[Ghosh et~al.(2007)Ghosh, Basu, and O’Mahony]{ghosh2007bayesian}
B.~Ghosh, B.~Basu, and M.~O’Mahony.
\newblock Bayesian time-series model for short-term traffic flow forecasting.
\newblock \emph{Journal of transportation engineering}, 133\penalty0
  (3):\penalty0 180--189, 2007.

\bibitem[Hammond et~al.(2011)Hammond, Vandergheynst, and
  Gribonval]{hammond2011wavelets}
D.~K. Hammond, P.~Vandergheynst, and R.~Gribonval.
\newblock Wavelets on graphs via spectral graph theory.
\newblock \emph{Applied and Computational Harmonic Analysis}, 30\penalty0
  (2):\penalty0 129--150, 2011.

\bibitem[Kipf and Welling(2016)]{kipf2016semi}
T.~N. Kipf and M.~Welling.
\newblock Semi-supervised classification with graph convolutional networks.
\newblock \emph{arXiv preprint arXiv:1609.02907}, 2016.

\bibitem[Kumar and Vanajakshi(2015)]{kumar2015short}
S.~V. Kumar and L.~Vanajakshi.
\newblock Short-term traffic flow prediction using seasonal arima model with
  limited input data.
\newblock \emph{European Transport Research Review}, 7\penalty0 (3):\penalty0
  1--9, 2015.

\bibitem[Li et~al.(2017)Li, Yu, Shahabi, and Liu]{li2017diffusion}
Y.~Li, R.~Yu, C.~Shahabi, and Y.~Liu.
\newblock Diffusion convolutional recurrent neural network: Data-driven traffic
  forecasting.
\newblock \emph{arXiv preprint arXiv:1707.01926}, 2017.

\bibitem[Li et~al.(2019)Li, Yu, Zhang, and Wang]{li2019bayesian}
Z.~Li, H.~Yu, G.~Zhang, and J.~Wang.
\newblock A bayesian vector autoregression-based data analytics approach to
  enable irregularly-spaced mixed-frequency traffic collision data imputation
  with missing values.
\newblock \emph{Transportation Research Part C: Emerging Technologies},
  108:\penalty0 302--319, 2019.

\bibitem[Liu et~al.(2020)Liu, Zhen, Li, Zhan, He, Du, and Lin]{liu2020dynamic}
L.~Liu, J.~Zhen, G.~Li, G.~Zhan, Z.~He, B.~Du, and L.~Lin.
\newblock Dynamic spatial-temporal representation learning for traffic flow
  prediction.
\newblock \emph{IEEE Transactions on Intelligent Transportation Systems}, 2020.

\bibitem[Ma et~al.(2015{\natexlab{a}})Ma, Tao, Wang, Yu, and Wang]{ma2015long}
X.~Ma, Z.~Tao, Y.~Wang, H.~Yu, and Y.~Wang.
\newblock Long short-term memory neural network for traffic speed prediction
  using remote microwave sensor data.
\newblock \emph{Transportation Research Part C: Emerging Technologies},
  54:\penalty0 187--197, 2015{\natexlab{a}}.

\bibitem[Ma et~al.(2015{\natexlab{b}})Ma, Yu, Wang, and Wang]{ma2015large}
X.~Ma, H.~Yu, Y.~Wang, and Y.~Wang.
\newblock Large-scale transportation network congestion evolution prediction
  using deep learning theory.
\newblock \emph{PloS one}, 10\penalty0 (3):\penalty0 e0119044,
  2015{\natexlab{b}}.

\bibitem[Okutani and Stephanedes(1984)]{okutani1984dynamic}
I.~Okutani and Y.~J. Stephanedes.
\newblock Dynamic prediction of traffic volume through kalman filtering theory.
\newblock \emph{Transportation Research Part B: Methodological}, 18\penalty0
  (1):\penalty0 1--11, 1984.

\bibitem[Oord et~al.(2016)Oord, Dieleman, Zen, Simonyan, Vinyals, Graves,
  Kalchbrenner, Senior, and Kavukcuoglu]{oord2016wavenet}
A.~v.~d. Oord, S.~Dieleman, H.~Zen, K.~Simonyan, O.~Vinyals, A.~Graves,
  N.~Kalchbrenner, A.~Senior, and K.~Kavukcuoglu.
\newblock Wavenet: A generative model for raw audio.
\newblock \emph{arXiv preprint arXiv:1609.03499}, 2016.

\bibitem[Sermanet and LeCun(2011)]{sermanet2011traffic}
P.~Sermanet and Y.~LeCun.
\newblock Traffic sign recognition with multi-scale convolutional networks.
\newblock In \emph{The 2011 International Joint Conference on Neural Networks},
  pages 2809--2813. IEEE, 2011.

\bibitem[Sun et~al.(2006)Sun, Zhang, and Yu]{sun2006bayesian}
S.~Sun, C.~Zhang, and G.~Yu.
\newblock A bayesian network approach to traffic flow forecasting.
\newblock \emph{IEEE Transactions on intelligent transportation systems},
  7\penalty0 (1):\penalty0 124--132, 2006.

\bibitem[Tremblay and Borgnat(2014)]{tremblay2014graph}
N.~Tremblay and P.~Borgnat.
\newblock Graph wavelets for multiscale community mining.
\newblock \emph{IEEE Transactions on Signal Processing}, 62\penalty0
  (20):\penalty0 5227--5239, 2014.

\bibitem[Van Der~Voort et~al.(1996)Van Der~Voort, Dougherty, and
  Watson]{van1996combining}
M.~Van Der~Voort, M.~Dougherty, and S.~Watson.
\newblock Combining kohonen maps with arima time series models to forecast
  traffic flow.
\newblock \emph{Transportation Research Part C: Emerging Technologies},
  4\penalty0 (5):\penalty0 307--318, 1996.

\bibitem[Vlahogianni et~al.(2014)Vlahogianni, Karlaftis, and
  Golias]{vlahogianni2014short}
E.~I. Vlahogianni, M.~G. Karlaftis, and J.~C. Golias.
\newblock Short-term traffic forecasting: Where we are and where we’re going.
\newblock \emph{Transportation Research Part C: Emerging Technologies},
  43:\penalty0 3--19, 2014.

\bibitem[Wang et~al.(2021)Wang, Zhang, Miao, and Yu]{wang2021mt}
S.~Wang, M.~Zhang, H.~Miao, and P.~S. Yu.
\newblock Mt-stnets: Multi-task spatial-temporal networks for multi-scale
  traffic prediction.
\newblock In \emph{Proceedings of the 2021 SIAM International Conference on
  Data Mining (SDM)}, pages 504--512. SIAM, 2021.

\bibitem[Williams(2001)]{williams2001multivariate}
B.~M. Williams.
\newblock Multivariate vehicular traffic flow prediction: evaluation of arimax
  modeling.
\newblock \emph{Transportation Research Record}, 1776\penalty0 (1):\penalty0
  194--200, 2001.

\bibitem[Wu et~al.(2019)Wu, Pan, Long, Jiang, and Zhang]{wu2019graph}
Z.~Wu, S.~Pan, G.~Long, J.~Jiang, and C.~Zhang.
\newblock Graph wavenet for deep spatial-temporal graph modeling.
\newblock \emph{arXiv preprint arXiv:1906.00121}, 2019.

\bibitem[Xu et~al.(2019)Xu, Shen, Cao, Qiu, and Cheng]{xu2019graph}
B.~Xu, H.~Shen, Q.~Cao, Y.~Qiu, and X.~Cheng.
\newblock Graph wavelet neural network.
\newblock \emph{arXiv preprint arXiv:1904.07785}, 2019.

\bibitem[Yu et~al.(2017{\natexlab{a}})Yu, Yin, and Zhu]{yu2017spatio}
B.~Yu, H.~Yin, and Z.~Zhu.
\newblock Spatio-temporal graph convolutional networks: A deep learning
  framework for traffic forecasting.
\newblock \emph{arXiv preprint arXiv:1709.04875}, 2017{\natexlab{a}}.

\bibitem[Yu et~al.(2020)Yu, Lee, and Sohn]{yu2020forecasting}
B.~Yu, Y.~Lee, and K.~Sohn.
\newblock Forecasting road traffic speeds by considering area-wide
  spatio-temporal dependencies based on a graph convolutional neural network
  (gcn).
\newblock \emph{Transportation research part C: emerging technologies},
  114:\penalty0 189--204, 2020.

\bibitem[Yu et~al.(2017{\natexlab{b}})Yu, Wu, Wang, Wang, and
  Ma]{yu2017spatiotemporal}
H.~Yu, Z.~Wu, S.~Wang, Y.~Wang, and X.~Ma.
\newblock Spatiotemporal recurrent convolutional networks for traffic
  prediction in transportation networks.
\newblock \emph{Sensors}, 17\penalty0 (7):\penalty0 1501, 2017{\natexlab{b}}.

\bibitem[Zang et~al.(2018)Zang, Ling, Wei, Tang, and Cheng]{zang2018long}
D.~Zang, J.~Ling, Z.~Wei, K.~Tang, and J.~Cheng.
\newblock Long-term traffic speed prediction based on multiscale
  spatio-temporal feature learning network.
\newblock \emph{IEEE Transactions on Intelligent Transportation Systems},
  20\penalty0 (10):\penalty0 3700--3709, 2018.

\bibitem[Zhang et~al.(2019{\natexlab{a}})Zhang, James, and
  Liu]{zhang2019spatial}
C.~Zhang, J.~James, and Y.~Liu.
\newblock Spatial-temporal graph attention networks: A deep learning approach
  for traffic forecasting.
\newblock \emph{IEEE Access}, 7:\penalty0 166246--166256, 2019{\natexlab{a}}.

\bibitem[Zhang and Li(2020)]{zhang2020ms}
M.~Zhang and Q.~Li.
\newblock Ms-gwnn: multi-scale graph wavelet neural network for breast cancer
  diagnosis.
\newblock \emph{arXiv preprint arXiv:2012.14619}, 2020.

\bibitem[Zhang et~al.(2019{\natexlab{b}})Zhang, Xie, Wang, Liu, and
  Wan]{zhang2019classifying}
R.~Zhang, P.~Xie, C.~Wang, G.~Liu, and S.~Wan.
\newblock Classifying transportation mode and speed from trajectory data via
  deep multi-scale learning.
\newblock \emph{Computer Networks}, 162:\penalty0 106861, 2019{\natexlab{b}}.

\bibitem[Zhao et~al.(2019)Zhao, Song, Zhang, Liu, Wang, Lin, Deng, and
  Li]{zhao2019t}
L.~Zhao, Y.~Song, C.~Zhang, Y.~Liu, P.~Wang, T.~Lin, M.~Deng, and H.~Li.
\newblock T-gcn: A temporal graph convolutional network for traffic prediction.
\newblock \emph{IEEE Transactions on Intelligent Transportation Systems},
  21\penalty0 (9):\penalty0 3848--3858, 2019.

\bibitem[Zhou et~al.(2020)Zhou, Cui, Hu, Zhang, Yang, Liu, Wang, Li, and
  Sun]{zhou2020graph}
J.~Zhou, G.~Cui, S.~Hu, Z.~Zhang, C.~Yang, Z.~Liu, L.~Wang, C.~Li, and M.~Sun.
\newblock Graph neural networks: A review of methods and applications.
\newblock \emph{AI Open}, 1:\penalty0 57--81, 2020.

\bibitem[Zhou et~al.(2017)Zhou, Qu, and Li]{zhou2017recurrent}
M.~Zhou, X.~Qu, and X.~Li.
\newblock A recurrent neural network based microscopic car following model to
  predict traffic oscillation.
\newblock \emph{Transportation research part C: emerging technologies},
  84:\penalty0 245--264, 2017.

\end{thebibliography}
\end{document}